\documentclass[review]{elsarticle}

\usepackage{lineno,hyperref}
\usepackage{amsmath}
\usepackage{amssymb}
\usepackage{csquotes}
\usepackage{tikz}
\usepackage{pgfplots}
\pgfplotsset{width=13cm,compat=1.9, height=5cm}
\usetikzlibrary{positioning,angles,quotes,arrows,shapes}
\modulolinenumbers[5]
\usepackage{caption}
\usepackage{subcaption}
\usepackage[ruled]{algorithm2e}
\usetikzlibrary{bayesnet}
\tikzset{>=latex}
\tikzstyle{plate caption} = [caption, node distance=0, inner sep=0pt,
below left=5pt and 0pt of #1.south]

\DeclareMathOperator*{\argmax}{arg\,max}
\newcommand{\E}{\mathbb{E}}

\DeclareMathOperator*{\dirichlet}{Dirichlet}
\DeclareMathOperator*{\categorical}{Categorical}
\DeclareMathOperator*{\TN}{TruncatedNormal}
\DeclareMathOperator*{\N}{Normal}
\DeclareMathOperator*{\TS}{TruncatedStudentT}
\DeclareMathOperator*{\betad}{Beta}
\DeclareMathOperator*{\G}{Gamma}
\DeclareMathOperator*{\ub}{ub}
\DeclareMathOperator*{\lb}{lb}

\journal{Journal of \LaTeX\ Templates}

\begin{document}

\begin{frontmatter}

\title{Bridging POMDPs and Bayesian decision making for robust maintenance planning under model uncertainty: An application to railway systems}


\author[ethaddress]{Giacomo Arcieri\corref{mycorrespondingauthor}}
\ead{giacomo.arcieri@ibk.baug.ethz.ch}

\author[ethaddress]{Cyprien Hoelzl}
\ead{hoelzl@ibk.baug.ethz.ch}

\author[sbbaddress]{Oliver Schwery}
\ead{oliver.schwery@sbb.ch}

\author[tumaddress]{Daniel Straub}
\ead{straub@tum.de}

\author[psuaddress]{Konstantinos G. Papakonstantinou}
\ead{kpapakon@psu.edu}

\author[ethaddress]{Eleni Chatzi}
\ead{chatzi@ibk.baug.ethz.ch}

\cortext[mycorrespondingauthor]{Corresponding author}

\address[ethaddress]{Institute of Structural Engineering, ETH Z{\"u}rich, 8093 Z{\"u}rich, Switzerland}
\address[sbbaddress]{Swiss Federal Railways SBB, 3000 Bern, Switzerland}
\address[tumaddress]{Engineering Risk Analysis Group, Technical University of Munich, 80333 Munich, Germany}
\address[psuaddress]{Dept. of Civil and Environmental Engineering, Pennsylvania State Univ., University Park, PA 16802, USA}

\begin{abstract}
Structural Health Monitoring (SHM) describes a process for inferring quantifiable metrics of structural condition, which can serve as input to support decisions on the operation and maintenance of infrastructure assets. Given the long lifespan of critical structures, this problem can be cast as a sequential decision making problem over prescribed horizons. Partially Observable Markov Decision Processes (POMDPs) offer a formal framework to solve the underlying optimal planning task. However, two issues can undermine the POMDP solutions. Firstly, the need for a model that can adequately describe the evolution of the structural condition under deterioration or corrective actions and, secondly, the non-trivial task of recovery of the observation process parameters from available monitoring data. Despite these potential challenges, the adopted POMDP models do not typically account for uncertainty on model parameters, leading to solutions which can be unrealistically confident. In this work, we address both key issues. We present a framework to estimate POMDP transition and observation model parameters directly from available data, via Markov Chain Monte Carlo (MCMC) sampling of a Hidden Markov Model (HMM) conditioned on actions. The MCMC inference estimates distributions of the involved model parameters. We then form and solve the POMDP problem by exploiting the inferred distributions, to derive solutions that are robust to model uncertainty. We successfully apply our approach on maintenance planning for railway track assets on the basis of a ``fractal value'' indicator, which is computed from actual railway monitoring data.
\end{abstract}


\end{frontmatter}

\linenumbers

\section{Introduction}
\label{sec:intro}
\nolinenumbers

Engineering infrastructures are subject to deterioration processes, which undermine a safe utilization and incur economic and environmental costs. Maintenance policies aim to extend the operating life-cycle, by seeking a trade--off between compromise in structural condition and the costs associated to repair and intervention actions. Structural Health Monitoring (SHM) contributes toward this goal by delivering data-driven indicators of structural condition, and/or by allowing to update and refine predictive models of operating engineered systems \cite{farrar2012structural}. The extracted information can support maintenance planning to achieve the long-term objectives of cost and risk minimization throughout the structural life-cycle. To this end, a probabilistic risk-based decision framework for SHM is outlined in \cite{hughes2022risk}. Linked to SHM is the concept of Value of Information (VoI) or Value of Structural Health Monitoring \cite{andriotis2021value, kamariotis2022value, giordano2022value, straub2017value}, which quantifies the cost benefits associated with adoption of monitoring tools. 

Cost efficient maintenance is crucial for effective management of extended infrastructure networks, as represented for instance in the case of railway systems. As a characteristic example, Switzerland’s railway network usage and load have increased by roughly $40\%$ and $70\%$, respectively, in the last 30 years, while the amount of traffic per km of track is the highest worldwide \cite{swissTAMP}. This increased backlog demand has led to higher life-cycle costs and an increase in disruptive events. However, infrastructure asset management has to obey budgetary, availability, and further constraints. As a result, new, efficient approaches for maintenance scheduling are needed to address modern challenges. In formalizing the approach to maintenance planning, it is possible to cast this as a sequential decision-making problem with a long horizon cost minimization objective \cite{ellis1995inspection}. Current decisions will bear an impact on the system's future condition, which - in absence of intervention - tends to stochastically evolve according to a degradation process. There is, however, significant uncertainty associated to the estimate of a system's condition, both at present and in the future. SHM offers a tool for more reliably tracking the system's state (condition), thus reducing the associated uncertainty. However, monitoring measurements come in the form of noise--corrupt information, which only approximate the actual structural state. This problem admits representation in the form of a Partially Observable Markov Decision Process (POMDP). The POMDP framework  utilizes the uncertain available information along with a (transition) model of the stochastic evolution of the system, to derive solutions with mathematically sound optimality properties \cite{papakonstantinou2014planning}. POMDPs have already been successfully implemented for solving optimal maintenance planning problems of  corroding reinforced concrete structures \cite{papakonstantinou2014planning2}, interstate highway pavements \cite{faddoul2013incorporating}, wind turbines \cite{memarzadeh2015optimal}, or deteriorating bridges \cite{papakonstantinou2018pomdp, schobi2016maintenance}. While POMDP solutions have long been limited to small-scale problems, it has recently been shown that the framework can be efficiently extended to more complex problems \cite{papakonstantinou2014planning2,papakonstantinou2016pomdp}. 

Nevertheless, there is currently scarce adoption and available literature of POMDP solutions for real-world applications. The framework requires knowledge of the stochastic transition dynamics of the structure as well as of the observation generating process. Such models are rarely available in the framework of infrastructure maintenance planning, but could be estimated from available data. However, the recovery of the involved transition dynamics and the associated observation model can be quite complex, while only scarce literature is available on best practices, as stated in \cite{papakonstantinou2014planning2}. As one of few examples, Papakonstantinou et al. \cite{papakonstantinou2014planning2} exploit a physical model, described in detail in \cite{papakonstantinou2013probabilistic}, in order to recover the state transition probability matrix for the deterioration process (i.e., action do-nothing, as explained in Section \ref{sec:pomdp_model}). However, the transition matrices for the repair actions, as well as the observation model, have not been derived from actual data. The authors themselves stress the need for further studies on recovering observation models and  transition models for maintenance actions.
In general, the majority of applications of POMDPs on infrastructure maintenance planning concern illustrative examples, often of simplified nature. Albeit these works are valuable, this reflects a lack of applications on real-world data, which would often necessitate inferring the transition dynamics relative to the deterioration process and maintenance actions, along with the associated observation model, entirely from data. This creates a gap between development of effective solution algorithms and their actual deployment to real-world applications. 

A main contribution of this work is to cover the aforementioned gap by formulating a framework that estimates directly and entirely from real-world data both the POMDP transition and observation models, via MCMC sampling from a Hidden Markov Model (HMM) conditioned on actions. We demonstrate the implementation of our approach based on a real-world problem of optimal maintenance planning for a railway network. Our inference technique can recover the full distributions of parameters, which represent all plausible values the model can assume under the available data. To this end, we exploit the ``fractal values'' indicator, collected across Switzerland's railway network and described in detail in Section \ref{sec:data}. While we focus on this specific application, the presented methods are general and applicable across a broader suite of problems. To the best of our knowledge, no other works demonstrate inference of the complete POMDP model entirely from real-world data.

A further critical point that prevents broad adoption in real-world applications, which is however only secondary to the inference of the complete POMDP model, is that POMDP solutions do not usually account
for epistemic uncertainty \cite{memarzadeh2016hierarchical}. Indeed, POMDP solutions are globally optimal for an assumed a-priori model structure, but this is unlikely to coincide with the actual environment (ground truth). As a result, POMDP solutions can be insufficiently robust against model uncertainty, often causing concerns when deployed on real-world applications. The work in \cite{ross2011bayesian} casts POMDPs into a fully Bayesian framework, but there is scarce literature on considering epistemic uncertainty in POMDP applications, with \cite{durango2002optimal, pozzi2017hidden, memarzadeh2015optimal} comprising few exceptions. Here, we build on these prior works to further bridge POMDPs and Bayesian decision making by considering model parameter distributions that are inferred by MCMC sampling. As a result, the computed POMDP solutions are not optimized for specific parameters but for all plausible values and are thus robust over model uncertainty. 

The remainder of this paper is organized as follows. The next section provides the POMDP theoretical background. Section \ref{sec:data} describes the ``fractal values'' indicator, namely the data used in this paper to recover transition and observation models, while Section \ref{sec:pomdp_model} explains how the problem of railway maintenance planning can be cast into the POMDP framework. Section \ref{sec:inference} illustrates the inference of the underlying transition and observation models,  Section \ref{sec:solution} presents the algorithms employed to derive policies that are robust to epistemic uncertainty and summarizes the results, and finally, Section \ref{sec:conclusion} concludes this work.

\section{Background and fundamentals}
\label{sec:bg}

\subsection{Markov Decision Process}

A Markov Decision Process (MDP) provides the mathematical framework for modelling a sequential decision making problem within a stochastic control setting. A MDP is defined by the tuple $\langle S, A,R,T,H,\gamma\rangle$, where:
\begin{itemize}
    \item $S$ is the finite set of states that the environment can assume.
    \item $A$ is the finite set of actions that the decision maker (or agent) can pick.
    \item $R:S \times A  \rightarrow\mathbb{R}$ is the reward function that assigns the reward $r_t=R(s_t, a_t)$ for assuming an action $a_t$ at state $s_t$.
    \item $T:S \times S \times A  \rightarrow [0,1]$ is the transition dynamics model that describes the probability $p(s_{t+1}|s_t,a_t)$ to transition to state $s_{t+1}$ if an action $a_t$ is taken at state $s_t$. Given that $S$ and $A$ are finite sets, there exists one transition matrix per possible action.
    \item $H$ is the considered horizon of the problem, which can be finite or infinite.
    \item $\gamma$ is the discount factor that discounts future rewards to obtain the present value. $\gamma\in [0,1]$ or $\gamma\in [0,1)$ depending on whether $H$ is assumed as finite or infinite, respectively. 
\end{itemize}
The objective of the MDP is to determine the optimal policy $\pi^*:S\rightarrow A$, which maps states to actions such that the expected sum of rewards is maximized:
\begin{equation}
    J(\pi^*)=\max_\pi\E\left[\sum_{t=0}^H\gamma^tr_t\right]
\end{equation}
where $r_t=R(s_t,\pi(s_t))$ and $\E\left[\cdot\right]$ is the expectation operator, which needs to be computed due to the stochastic nature of the transition dynamics.

\begin{figure}[htb]
\begin{tikzpicture}[auto,node distance=8mm,>=latex,font=\small]
\tikzset{
    >=stealth',
    node distance=1.5cm,
    state/.style={minimum size=50pt,font=\small,circle,draw},
    dots/.style={state,draw=none},
    edge/.style={->},
    trans/.style={font=\footnotesize,above=2mm},
    reflexive/.style={out=120,in=60,looseness=5,relative},
    squared/.style={rectangle, draw=black, fill=white, thick, minimum size=8mm},
    round/.style={circle, draw=black, fill=white, thick, minimum size=8mm},
    decision/.style={diamond, draw=black, fill=white, thick, minimum size=8mm},
    minimum size=8mm,inner sep=0pt
  }
    
    \node[round, fill=gray!25] (s0) {$s_t$};
    \node [dots]  (d0)  [left=20mm of s0] {$\cdots$};
    \node[squared] (a0) [below right=10mm and 5mm of s0] {$a_t$};
    \node[decision] (r0) [above right=10mm and 5mm of s0] {$r_t$};
    \node[round, fill=gray!25 ,right=20mm of s0] (s1) {$s_{t+1}$};
    \node[squared] (a1) [below right=10mm and 5mm of s1] {$a_{t+1}$};
    \node[decision] (r1) [above right=10mm and 5mm of s1] {$r_{t+1}$};
    \node [dots]  (d1)  [right=20mm of s1] {$\cdots$};
    
    \draw[->] (d0)to node[above right=-1.5mm and 0.mm]{ $T$}(s0);
    \draw[->] (s0)to node[above right=-1.5mm and 0.mm]{ $T$}(s1);
    \draw[->] (s0)to node[above=-1mm]{ $\pi$}(a0);
     \draw[->] (s0)--(r0);
    \draw[->] (a0)--(s1);
     \draw[->] (a0)to node[above left=3.5mm and -1mm]{ $R$}(r0);
    \draw[->] (s1)to node[above right=-1.5mm and 0.mm]{ $T$}(d1);
    \draw[->] (s1)to node[above=-1mm]{ $\pi$}(a1);
    \draw[->] (s1)--(r1);
     \draw[->] (a1)to node[above left=3.5mm and -1mm]{ $R$}(r1);
    \draw[->] (a1)--(d1);

\end{tikzpicture}
\caption{Probabilistic graphical model of a MDP.}
\label{fig:mdp}
\end{figure}
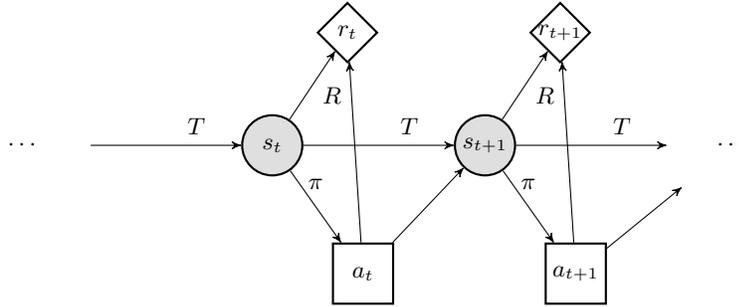

An MDP can be represented as a special case of influence diagrams \cite{morato2020optimal, luque2019risk}; which form a class of probabilistic graphical models. Figure \ref{fig:mdp} illustrates the graphical model for a general MDP. Circles, rectangles and diamonds correspond to random, decision and utility variables, respectively \cite{koller2009probabilistic}. Shaded shapes denote observed variables, while edges indicate dependencies among variables.

An MDP is assumed to satisfy the Markov property, which states that the current state $s_t$ contains all information on past agent-environment interactions that bear an effect on the future state \cite{puterman2014markov}, i.e., $p(s_{t+1}|s_t,a_t,\dots,s_0,a_0)=p(s_{t+1}|s_t,a_t)$. If a process does not satisfy the Markov property, the problem may still be modeled as an MDP by state augmentation \cite{papakonstantinou2014planning}. In such an approach, the state vector $s_t$ is augmented to further include previous information so that the Markov property is satisfied. Likewise, time can be encoded in the state, allowing to model non-stationary problems and to transform finite horizon problems into infinite ones.

The MDP problem can be solved via Dynamic Programming techniques \cite{bertsekas2012dynamic} and the introduction of the value function $V^\pi:S\rightarrow \mathbb{R}$, which represents the expected sum of rewards of policy $\pi$ from a certain state. The optimal policy $\pi^*$ can be computed through Bellman's optimality equation:

\begin{equation}\label{eq:bellman_opt}
    V^{\pi^*}_n(s_t)=\max_{a_t\in A}\left[R(s_t,a_t)+\gamma\sum_{s_{t+1}\in S}p(s_{t+1}|s_t,a_t)V^{\pi^*}_{n-1}(s_{t+1})\right]
\end{equation}
Equation \ref{eq:bellman_opt} can be solved with the value iteration algorithm \cite{bellman1966dynamic}. For the finite horizon problem, $n$ represents the remaining steps for reaching horizon $H$, i.e., the algorithm operates backwards, initiating at the last time step and identifying the optimal actions for all preceding steps. For the discounted infinite horizon problem, $\gamma$ must be strictly less than one in order for the computed state-values in Equation \ref{eq:bellman_opt} to be finite, all state-values are initialized, commonly at zero, and Equation \ref{eq:bellman_opt} is iterated until some convergence threshold is reached. Thus, in the infinite horizon case, $n$ represents the iteration step of the algorithm. If the transition probabilities in Equation \ref{eq:bellman_opt} are not known, state-values (and, hence, an optimal policy) can be learned with reinforcement learning via temporal difference methods \cite{sutton1988learning}.

Bellman's equation can alternatively be written in terms of the Q-value function \cite{sutton2018reinforcement}:
\begin{equation}\label{eq:q_value}
    Q^{\pi^*}_n(s_t, a_t)=R(s_t,a_t)+\gamma\sum_{s_{t+1}\in S}p(s_{t+1}|s_t,a_t)V^{\pi^*}_{n-1}(s_{t+1})
\end{equation}
which outputs the state-value for taking action $a_t$ at state $s_t$ and then following the optimal policy $\pi^*$. Hence, it represents a one-time deviation from the considered policy. The state-value function can then be obtained by choosing the action that maximizes the Q-value function:
\begin{equation}\label{eq:bellman_opt2}
    V^{\pi^*}_n(s_t)=\max_{a_t\in A}\left[Q^{\pi^*}_n(s_t, a_t)\right]
\end{equation}

\subsection{Partially Observable Markov Decision Process}
\label{sec:bg_pomdp}

A POMDP extends the MDP framework by incorporating uncertainty into the observations. The states are now hidden variables, which generate observations that provide partial and/or noisy information about the actual state of the system. A POMDP is thus defined by the tuple $\langle S,A,Z,R,T,O,b_0,H,\gamma\rangle$, where the newly introduced variables are:
\begin{itemize}
    \item $Z$ is the set of possible observations.
    \item $O:S \times A \times Z  \rightarrow \mathbb{R}$ is the observation generating process that defines the emission probability $p(z_t|s_t,a_{t-1})$, namely the likelihood to observe $z_t$ if the system is at state $s_t$ and action $a_{t-1}$ was taken. If $Z$ is finite, then the output is defined over the interval $[0,1]$.
    \item $b_0$ is the initial belief on the state of the system $s_0$, with the belief variable defined in what follows.
\end{itemize}

\begin{figure}[htb]
\begin{tikzpicture}[auto,node distance=8mm,>=latex,font=\small]
\tikzset{
    >=stealth',
    node distance=1.5cm,
    state/.style={minimum size=50pt,font=\small,circle,draw},
    dots/.style={state,draw=none},
    edge/.style={->},
    trans/.style={font=\footnotesize,above=2mm},
    reflexive/.style={out=120,in=60,looseness=5,relative},
    squared/.style={rectangle, draw=black, fill=white, thick, minimum size=8mm},
    round/.style={circle, draw=black, fill=white, thick, minimum size=8mm},
    round_hidden/.style={circle, draw=black, fill=gray!25, thick, minimum size=8mm},
    decision/.style={diamond, draw=black, fill=white, thick, minimum size=8mm},
    minimum size=8mm,inner sep=0pt
  }
    
    \node[round] (b0) {$b_t$};
    \node[round_hidden] (o0) [above=15mm of b0] {$z_t$};
    \node[round] (s0) [above=15mm of o0] {$s_t$};
    \node [dots]  (d0)  [left=20mm of b0] {};
    \node [dots]  (ds0)  [left=20mm of s0] {};
    \node [dots]  (do0)  [left=20mm of o0] {$\cdots$};
    \node[squared] (a0) [below right=5mm and 5mm of s0] {$a_t$};
    \node[decision] (r0) [above right=10mm and 5mm of s0] {$r_t$};
    \node[round,right=20mm of b0] (b1) {$b_{t+1}$};
    \node[round_hidden] (o1) [above=15mm of b1] {$z_{t+1}$};
    \node[round] (s1) [above=15mm of o1] {$s_{t+1}$};
    \node[squared] (a1) [below right=5mm and 5mm of s1] {$a_{t+1}$};
    \node[decision] (r1) [above right=10mm and 5mm of s1] {$r_{t+1}$};
    \node [dots]  (d1)  [right=20mm of b1] {};
    \node [dots]  (ds1)  [right=20mm of s1] {};
    \node [dots]  (do1)  [right=20mm of o1] {$\cdots$};
    
    \draw[->] (d0)--(b0);
    \draw[->] (ds0)to node[above right=-1.5mm and 0.mm]{ $T$}(s0);
    \draw[->] (b0)--(b1);
    \draw[->] (s0)to node[above right=-1.5mm and 0.mm]{ $T$}(s1);
    \draw[->] (s0)to node[below left=0mm and -1mm]{ $O$}(o0);
    \draw[->] (o0)--(b0);
    \draw[->] (b0)to node[above right=-5mm and -3mm]{ $\pi$}(a0);
     \draw[->] (s0)--(r0);
    \draw[->] (a0)--(b1);
    \draw[->] (a0)--(s1);
    \draw[->] (a0)--(o1);
     \draw[->] (a0)to node[above left=3.5mm and -1mm]{ $R$}(r0);
    \draw[->] (b1)--(d1);
    \draw[->] (s1)to node[above right=-1.5mm and 0.mm]{ $T$}(ds1);
    \draw[->] (s1)to node[below left=0mm and -1mm]{ $O$}(o1);
    \draw[->] (o1)--(b1);
    \draw[->] (b1)to node[above right=-5mm and -3mm]{ $\pi$}(a1);
    \draw[->] (s1)--(r1);
    \draw[->] (a1)to node[above left=3.5mm and -1mm]{ $R$}(r1);
    \draw[->] (a1)--(d1);
    \draw[->] (a1)--(ds1);
    \draw[->] (a1)--(do1);

\end{tikzpicture}
\caption{Probabilistic graphical model of a POMDP.}
\label{fig:pomdp}
\end{figure}
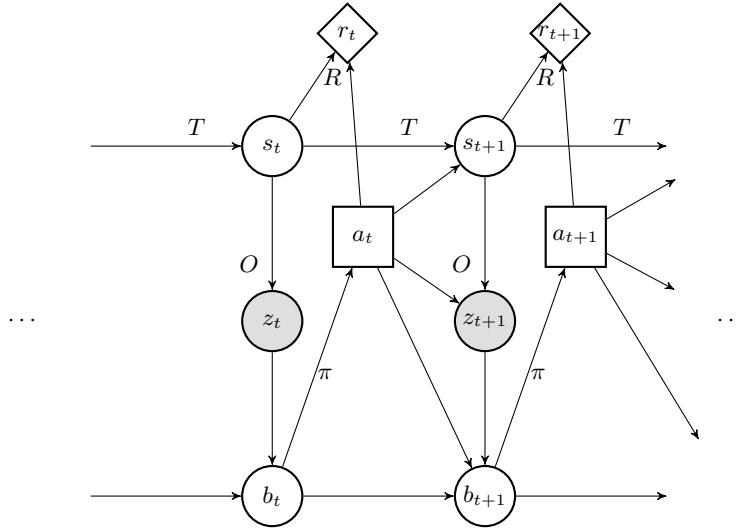

Given the partial information that the observations provide, the agent should take actions based on the full observation history, which would violate the Markov property. As such, a new variable is introduced in the POMDP setting: the belief state $\mathbf{b}$. The belief is a probability distribution over $S$, which maps the discrete finite set of states into a continuous $|S|-1$ dimensional simplex \cite{papakonstantinou2014planning}. The belief over the state of the system is updated every time the agent receives a new observation according to Bayes' rule:
\begin{equation}\label{eq:belief}
    b(s_{t+1})=\frac{p(z_{t+1}|s_{t+1},a_t)}{p(z_{t+1}|\mathbf{b},a_t)}\sum_{s_t\in S}p(s_{t+1}|s_t,a_t)b(s_t)
\end{equation}
where the denominator is the usual normalizing factor:
\begin{equation}
    p(z_{t+1}|\mathbf{b},a_t)=\sum_{s_{t+1}\in S}p(z_{t+1}|s_{t+1},a_t)\sum_{s_t\in S}p(s_{t+1}|s_t,a_t)b(s_t)
\end{equation}
The belief over the state of the system at time $t$ offers sufficient statistics of the full history of actions and observations, namely it provides the decision maker with the same amount of information. The decision maker can then follow a policy $\pi(\mathbf{b})$, which depends on the computed belief, and the POMDP framework thus satisfies the Markov property. The probabilistic graphical model of the POMDP is provided in Figure \ref{fig:pomdp}, whereby state variables are no longer observed, but are hidden variables.

The previously defined Bellman equation changes accordingly to:

\begin{equation}\label{eq:bellman_opt_pomdp}
    V^{\pi^*}_n(\mathbf{b})=\max_{a_t\in A}\left[\sum_{s_{t}\in S}b(s_t)R(s_t,a_t)+\gamma\sum_{z_{t+1}\in Z}p(z_{t+1}|\mathbf{b},a_t)V^{\pi^*}_{n-1}(\mathbf{b}^\prime)\right]
\end{equation}
where $\mathbf{b}^\prime$ is the updated belief, which is computed according to Equation \ref{eq:belief}. 

Solving a POMDP is thus equivalent to solving a continuous state MDP defined over the belief space. While it is still possible to provide optimality convergence properties of the value iteration algorithm thanks to the piecewise linear convex property \cite{papakonstantinou2014planning}, the exact solution is generally intractable except for very low-dimensional problems. As such, in the literature POMDP solution methods have been relying on approximations, such as \cite{parr1995approximating}. The advent of point-based value iteration algorithms allowed to efficiently solve large scale POMDP problems with good approximation, although they generally require $S$, $A$, and $Z$ to be finite. An introduction to these methods is provided in \cite{spaan2005perseus, papakonstantinou2018pomdp}.

\subsection{Bayesian Decision Making}
\label{sec:bayesian_decision_making}
In the previous sections, we introduced the transition dynamics $T$ and the observation generating process $O$. These models generally depend on some parameters $\mathbf{\theta}$. In existing literature, these are typically treated as fixed. However, in many applications these parameters can be subject to uncertainty, often due to the limited amount of data used for learning, leading to epistemic uncertainty. To tackle this, a number of works \cite{ross2011bayesian, memarzadeh2015optimal,memarzadeh2016hierarchical} cast the sequential decision-making problem of a POMDP into a fully Bayesian framework. Indeed, while the POMDP framework is inherently Bayesian, due to the update of the belief variable through Bayes theorem, the scheme is not generally treated as a fully Bayesian framework, since POMDP parameters are not considered as random variables $p(\mathbf{\theta})$, thus failing to incorporate model uncertainty into the solution.

In Bayesian decision theory \cite{berger2013statistical} the concept of utility function $U(\mathbf{\theta}, a)$ is introduced, which maps possible outcomes to their utility given the parameters $\mathbf{\theta}$ and some decision $a$. The Bayesian optimal action is the one which maximizes the expected utility:
\begin{equation}
    a^*=\argmax_{a\in A}\E_{\mathbf{\theta}\sim p(\mathbf{\theta})}\left[U(\mathbf{\theta}, a)\right]
\end{equation}

In the MDP framework, the concept of utility is associated with the Q-values. We denote $Q^{\pi}_\mathbf{\theta}(s, a)$ as the Q-value for action $a$ when the model parameters are $\mathbf{\theta}\sim p(\mathbf{\theta})$. In this fully Bayesian context, the optimal action thus maximizes the expected Q-value function over the model parameter distribution \cite{memarzadeh2015optimal}:
\begin{equation}\label{eq:bayes_optimal_action_q}
    a^*=\argmax_{a\in A}\E_{\mathbf{\theta}\sim p(\mathbf{\theta})}\left[Q^{\pi^*}_\mathbf{\theta}(s, a)\right]
\end{equation}
In this setting, the optimal policy may be sub-optimal for a specific value $\mathbf{\theta}$, while maximizing the expected value with respect to the entire model parameter distribution. As a result, the policy is robust over epistemic uncertainty.

\section{Data description} 
\label{sec:data}

\begin{figure}[h]
\begin{tikzpicture}
  \node (img)  {\includegraphics[width=10cm]{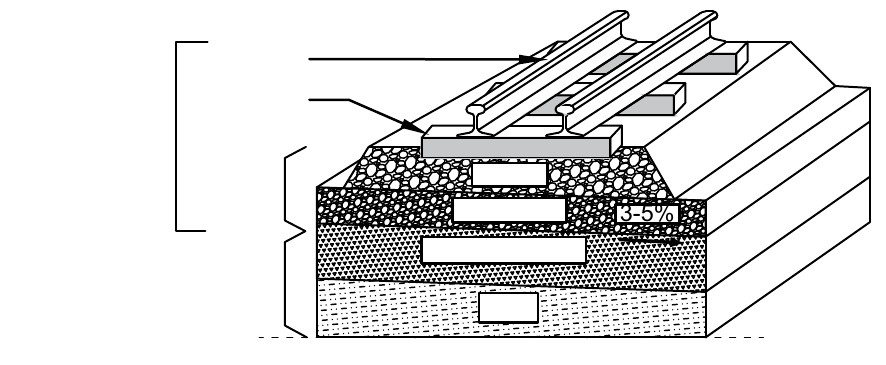}};
  \node[below=of img, node distance=0cm,scale=0.75, anchor=center,xshift=0.98cm, yshift=4.36cm] {ballast};
  \node[below=of img, node distance=0cm,scale=0.75, anchor=center,xshift=0.98cm, yshift=3.84cm] {subballast};
  \node[below=of img, node distance=0cm,scale=0.75, anchor=center,xshift=0.937cm, yshift=3.2cm] {formation layer};
  \node[below=of img, node distance=0cm,scale=0.75, anchor=center,xshift=0.98cm, yshift=2.38cm] {base};
  \node[below=of img, node distance=0cm,scale=0.75, anchor=center,xshift=-3.2cm, yshift=4.1cm] {track bed};
  \node[below=of img, node distance=0cm,scale=0.75, anchor=center,xshift=-3.45cm, yshift=2.8cm] {substructure};
  \node[below=of img, node distance=0cm,scale=0.75, anchor=center,xshift=-5.25cm, yshift=4.9cm] {superstructure};
  \node[below=of img, node distance=0cm,scale=0.75, anchor=center,xshift=-2.5cm, yshift=6.15cm] {rail};
  \node[below=of img, node distance=0cm,scale=0.75, anchor=center,xshift=-2.75cm, yshift=5.5cm] {sleeper};
 \end{tikzpicture}
\caption{Structure of the railway track.}
\label{fig:structuretrack}
\end{figure}
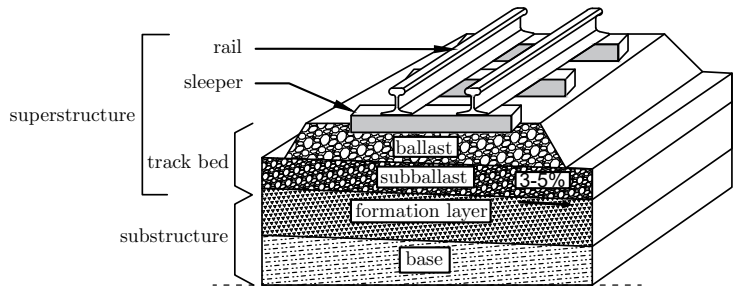

Although our suggested techniques are generally applicable, the focus application in this work is related to maintenance planning for railway track infrastructure. The latter forms an assembly of  multiple components (rails, sleepers, ballast, switches, etc.), as illustrated in Figure \ref{fig:structuretrack}, which are exposed to harsh environments and high loads, leading to accelerated degradation. The durability of the railway track, as well as its renewal costs are strongly dependent on the condition of certain components, such as the substructure. The substructure plays an essential role in the degradation process of the track, as the substructure material sustains cyclic loading from the superstructure, acts as a filter that blocks the uprising of fine particles into the ballast, and facilitates water drainage. A weakened substructure will typically result in distortions of the track geometry. Tamping, a maintenance action involving the usage of compacting devices to pack the ballast under the railway track, is often applied when the substructure condition is deemed moderately deteriorated. When only the superstructure is degraded (ballast fouling) the preferred maintenance measures are ballast cleaning or replacement. If the substructure is in poor condition (intrusion of clay or mud, water clogging, etc.), tamping or superstructure maintenance can only provide a short-term remedy, leaving replacement of the superstructure and substructure as the most appropriate long-term solution. 
Clearly, the optimization of maintenance decisions for such critical infrastructure components would benefit from information that is additional to scheduled inspection. Such additional information can be delivered from monitoring data derived by diagnostic vehicles. In this work, we specifically exploit the \textit{fractal values}, a substructure condition indicator derived from diagnostic vehicle measurements to guide decisions for substructure renewal. 

Such diagnostic vehicles form part of modern practice in the management of infrastructure assets. In the domain of railway infrastructure predictive or reactive maintenance and renewal decisions are increasingly guided by data-supported decision tools, such as the SwissTamp platform of the Swiss Federal Railways \cite{swissTAMP}.
In the railways context, data driven decision-making involves the following data types: infrastructure inventory information, maintenance data, usage frequency, loads, and periodic inspection data. Periodic inspection is carried out by means of diagnostic measurement vehicles that are equipped with a multitude of sensors (cameras, accelerometers, laser-distometers, etc). Amongst the diverse portfolio of collected information, the track geometry measurements, in particular, deliver condition indicators that are readily exploited for the network-wide estimation of the ballast and substructure condition \cite{HoelzlReview2021}. 

A specific set of such condition indicators are the so-called fractal values, which are derived from the longitudinal level measurement. The longitudinal level represents the vertical smoothness of the rail and is measured via a diagnostic vehicle as the deviation of the running surface of the rail from the smoothed vertical position \cite{wang2021study}.
Fractal values are the outcome of fractal analysis; a method that was originally developed as a way to approximate the length of the British coastline on the basis of Euclidean geometry principles. 
The fractal dimension, first termed by Benoit Mandelbrot \cite{Mandelbrot1967}, corresponds to the ratio between the change in the details in a pattern with respect to the change in the measurement scale. For railway tracks the fractal dimension corresponds to the degree of ``roughness" at varying wavelength scales. For the interested reader, the detailed steps of the fractal value computation are reported in Algorithm~\ref{f:algoFraq} of the Appendix, which was devised by Matthias Landgraf \cite{Landgraf2019}. The fractal values, first applied to railway condition assessment by Landgraf \cite{Landgraf2016}, are now used in practice by the Austrian and Swiss railways to detect ballast and substructure damage \cite{Landgraf2019}. Mid-wave (3-25~m) fractal values have been shown to have a higher correlation to ballast degradation, while long-wave (25-70~m) fractal values are more related to substructure damages \cite{HoelzlIABMAS2021}. 

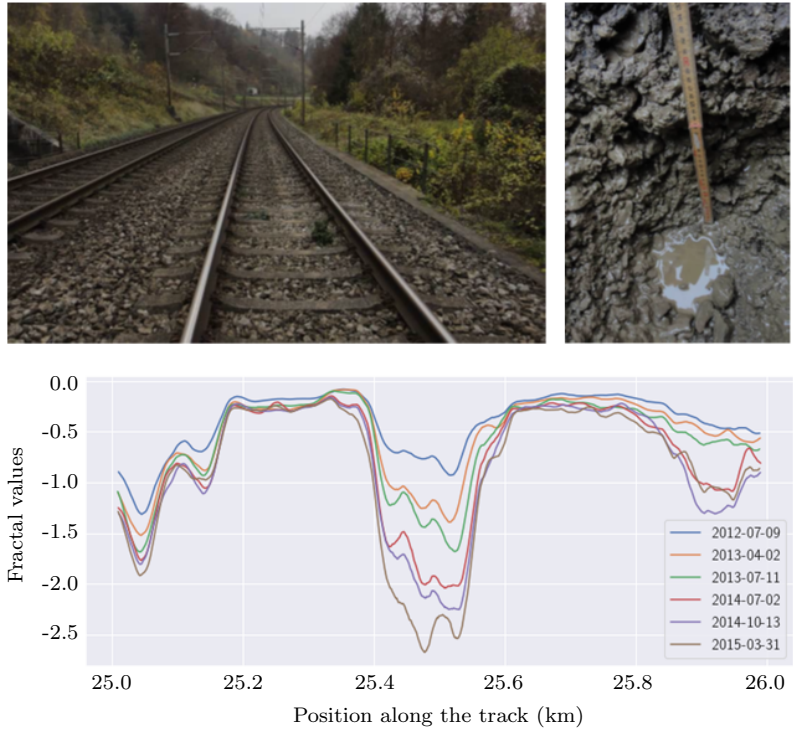
\begin{figure}[htb]
\begin{tikzpicture}
  \node (img)  {
  \includegraphics[width=0.9\textwidth]{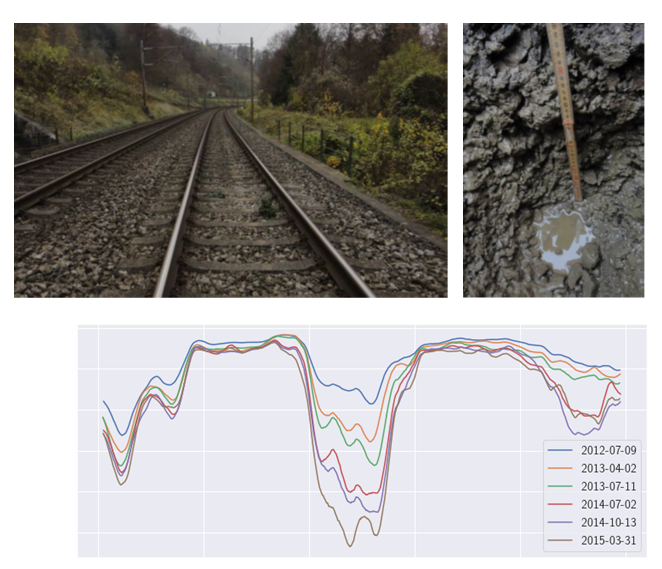}};
  \node[left=of img, node distance=0cm, rotate=90, anchor=center,xshift=-2.5cm, yshift=-1.45cm] {\footnotesize Fractal values};
  \node[left=of img, node distance=0cm, anchor=center,xshift=2.05cm, yshift=-0.74cm] {\footnotesize \ 0.0};
  \node[left=of img, node distance=0cm, anchor=center,xshift=2.05cm, yshift=-1.39cm] {\footnotesize -0.5};
  \node[left=of img, node distance=0cm, anchor=center,xshift=2.05cm, yshift=-2.05cm] {\footnotesize -1.0};
  \node[left=of img, node distance=0cm, anchor=center,xshift=2.05cm, yshift=-2.72cm] {\footnotesize -1.5};
  \node[left=of img, node distance=0cm, anchor=center,xshift=2.05cm, yshift=-3.4cm] {\footnotesize -2.0};
  \node[left=of img, node distance=0cm, anchor=center,xshift=2.05cm, yshift=-4.06cm] {\footnotesize -2.5};
  \node[below=of img, node distance=0cm,xshift=0.5cm,  yshift=0.9cm] {\footnotesize Position along the track (km)};
  \node[below=of img, node distance=0cm,xshift=-3.85cm,  yshift=1.3cm] {\footnotesize 25.0};
  \node[below=of img, node distance=0cm,xshift=-2.1cm,  yshift=1.3cm] {\footnotesize 25.2};
  \node[below=of img, node distance=0cm,xshift=-0.35cm,  yshift=1.3cm] {\footnotesize 25.4};
  \node[below=of img, node distance=0cm,xshift=1.36cm,  yshift=1.3cm] {\footnotesize 25.6};
  \node[below=of img, node distance=0cm,xshift=3.08cm,  yshift=1.3cm] {\footnotesize 25.8};
  \node[below=of img, node distance=0cm,xshift=4.84cm,  yshift=1.3cm] {\footnotesize 26.0};
\end{tikzpicture}
\caption{A highly deteriorated track from an inspection in 2014. The upper left image shows an overview of the track at the deteriorated location. The right image shows a severely degraded portion of the track at km 25.5, with presence of fouling, clay, and water intrusion. The bottom left image shows the associated long-wave fractal values. The decreasing values around km 25.5 indicate the deterioration of the area over time.}
\label{fig:deteriorated_track}
\end{figure}

A visual example is offered in Figure \ref{fig:deteriorated_track}, which displays a highly deteriorated portion of a track in 2014. The area shows presence of clay, fine material (fouling), and water intrusion, which represent characteristic problems of ballast and substructure damages. The figure also reports fractal value data that has been collected over the same area from 2012 and 2015. As a result of the deterioration of the track, fractal values decrease over time. In the damaged area (km 25.5 in figure), the fractal values have dropped considerably in the examined time-frame, suggesting the severe degradation confirmed by the inspection.

In this work, we use actual track geometry measurements, carried out by the SBB (the Swiss Federal Railways) between 2008 and 2018, across Switzerland's railway network, for tracks whose superstructure or substructure were subsequently maintained in 2019 \cite{HoelzlIABMAS2021}. The track geometry measurements were collected at least twice a year for the tracks under investigation. The fractal values are computed every 2.5m from the measured longitudinal level. The performed maintenance actions have been additionally logged for the analysed tracks. These logs contain information on the maintenance, repair, or renewal actions taken on a section of the network at a specific date.
We propose herein a unique POMDP scheme, which relies on diagnostic vehicle measurements of long-wave fractal values to predict an optimal maintenance policy.

\section{POMDP modeling}
\label{sec:pomdp_model}
Within our  application of railway maintenance planning, the POMDP problem is defined by the following variables:
\begin{itemize}
    \item \textit{Hidden states}, which represent the health condition of the track. We assume 4 hidden states: $s_0$, $s_1$, $s_2$, and $s_3$, which can be seen as \textit{very good}, \textit{good}, \textit{bad}, and \textit{very bad} track conditions, respectively. The choice of the number of hidden states is eventually arbitrary, since ground truth is not available. However, we adopted a pragmatic approach for determining the dimension of hidden states, by assuming this as a hyperparameter and repeating the inference of the model presented in the next section for 3, 4 and 5 hidden states. The model with 4 hidden states yielded improved convergence and better-defined distributions. In addition, 4 discrete condition states are assumed in similar works \cite{papakonstantinou2014planning2}.
    \item \textit{Actions}, represented by the possible maintenance actions. We focus on 3 possible actions among the ones recorded in our available data. $a_0$ represents the do-nothing action, i.e., the agent chooses not to take any maintenance at this decision step. The effect of $a_0$ is governed by the degradation process. $a_1$ is a low cost ``tamping'' action, which is often conducted as part of standard ballasted track maintenance. Tamping vehicles are commonly used to restore the geometry of ballasted tracks in a nearly automatic fashion \cite{Audley2013}.  Finally, $a_2$ is a more costly repair action, which involves the renewal of the substructure plus maintenance similar to $a_1$. In the offered case study, we demonstrate how the effects of $a_1$ and $a_2$ can be learned on the basis of the efficacy of these repair actions.
    \item \textit{Observations}, defined by the fractal values. The decision maker forms a belief over the state condition of the track, on the basis of the fractal values indicator, and makes a decision to follow one of the aforementioned actions. Fractal values comprise negative, continuous values which tend to decrease if no maintenance action is taken. The fractal values observed in our actual data, over the averaged observation lengths, reflect a clear negative trend, which motivate an attempt to model these observations as dependent on the previous value in order to ensure temporal coherence, introducing an autoregressive property among observations. Practical examples of the need for this property are given in the next section. 
    \item \textit{(Negative) rewards}, representing costs associated with actions and states. Typically, the costs of actions can be defined by the infrastructure operator. Quantifying the cost of different states is a far more difficult task. It should include costs and economic risks such as the deterioration of service due to imperfect track conditions, delays, environmental costs, working accidents or derailing risks. Hence, these costs are hard to quantify but crucial to justify maintenance expenses. We discussed both classes of costs with our SBB partners and report them in Table \ref{tab:costs} in general cost units, although only cost ratios matter for the solution of the problem. The action do-nothing does not have any cost. Action $a_1$ costs 50 (units) regardless of the condition of the track. The cost of the renewal part of action $a_2$ varies from 2,000 to 4,000 units depending on the condition of the structure, plus 50 units due to the tamping action.
\end{itemize}

\begin{table}[!h]
\caption{Costs of the POMDP model.}
    \centering
    \begin{tabular}{ccccc}
        \multicolumn{1}{c}{\bf State condition}  &\multicolumn{1}{c}{\bf $s_0$}  &\multicolumn{1}{c}{\bf $s_1$} &\multicolumn{1}{c}{\bf $s_2$}
        &\multicolumn{1}{c}{\bf $s_3$}\\
        \hline
        \bf Maintenance action\\
        $a_0$ & $0$   & $0$   & $0$ & $0$ \\
        $a_1$ & $-50$ & $-50$ & $-50$ & $-50$ \\
        $a_2$ & $-2,050$ & $-2,710$ & $-3,370$ & $-4,050$ \\
        \textbf{Condition cost} & $-100$ & $-200$ & $-1,000$ & $-8,000$\\
    \end{tabular}
    \label{tab:costs}
\end{table}

\begin{figure}[htb]
\begin{tikzpicture}[auto,node distance=8mm,>=latex,font=\small]
\tikzset{
    >=stealth',
    node distance=1.5cm,
    state/.style={minimum size=50pt,font=\small,circle,draw},
    dots/.style={state,draw=none},
    edge/.style={->},
    trans/.style={font=\footnotesize,above=2mm},
    reflexive/.style={out=120,in=60,looseness=5,relative},
    squared/.style={rectangle, draw=black, fill=white, thick, minimum size=8mm},
    round/.style={circle, draw=black, fill=white, thick, minimum size=8mm},
    round_hidden/.style={circle, draw=black, fill=black!10, thick, minimum size=8mm},
    decision/.style={diamond, draw=black, fill=white, thick, minimum size=8mm},
    minimum size=8mm,inner sep=0pt
  }
    
    \node[round] (b0) {$b_t$};
    \node[round_hidden] (o0) [above=15mm of b0] {$z_t$};
    \node[round] (s0) [above=15mm of o0] {$s_t$};
    \node [dots]  (d0)  [left=20mm of b0] {};
    \node [dots]  (ds0)  [left=20mm of s0] {};
    \node [dots]  (do0)  [left=20mm of o0] {$\cdots$};
    \node[squared] (a0) [below right=5mm and 5mm of s0] {$a_t$};
    \node[decision] (r0) [above right=10mm and 5mm of s0] {$r_t$};
    \node[round,right=20mm of b0] (b1) {$b_{t+1}$};
    \node[round_hidden] (o1) [above=15mm of b1] {$z_{t+1}$};
    \node[round] (s1) [above=15mm of o1] {$s_{t+1}$};
    \node[squared] (a1) [below right=5mm and 5mm of s1] {$a_{t+1}$};
    \node[decision] (r1) [above right=10mm and 5mm of s1] {$r_{t+1}$};
    \node [dots]  (d1)  [right=20mm of b1] {};
    \node [dots]  (ds1)  [right=20mm of s1] {};
    \node [dots]  (do1)  [right=20mm of o1] {$\cdots$};
    
    \draw[->] (d0)--(b0);
    \draw[->] (ds0)to node[above right=-1.5mm and 0.mm]{ $T$}(s0);
    \draw[->] (b0)--(b1);
    \draw[->] (s0)to node[above right=-1.5mm and 0.mm]{ $T$}(s1);
    \draw[->] (s0)to node[below left=0mm and -1mm]{ $O$}(o0);
    \draw[->] (do0)--(o0);
    \draw[->] (o0)--(b0);
    \draw[->] (o0)--(o1);
    \draw[->] (b0)to node[above right=-5mm and -3mm]{ $\pi$}(a0);
     \draw[->] (s0)--(r0);
    \draw[->] (a0)--(b1);
    \draw[->] (a0)--(s1);
    \draw[->] (a0)--(o1);
     \draw[->] (a0)to node[above left=3.5mm and -1mm]{ $R$}(r0);
    \draw[->] (b1)--(d1);
    \draw[->] (s1)to node[above right=-1.5mm and 0.mm]{ $T$}(ds1);
    \draw[->] (s1)to node[below left=0mm and -1mm]{ $O$}(o1);
    \draw[->] (o1)--(b1);
    \draw[->] (o1)--(do1);
    \draw[->] (b1)to node[above right=-5mm and -3mm]{ $\pi$}(a1);
    \draw[->] (s1)--(r1);
    \draw[->] (a1)to node[above left=3.5mm and -1mm]{ $R$}(r1);
    \draw[->] (a1)--(d1);
    \draw[->] (a1)--(ds1);
    \draw[->] (a1)--(do1);

\end{tikzpicture}
\caption{Probabilistic graphical model of the considered POMDP. The dependency among observations is displayed with an additional arrow among these variables.}
\label{fig:pomdp_railway}
\end{figure}
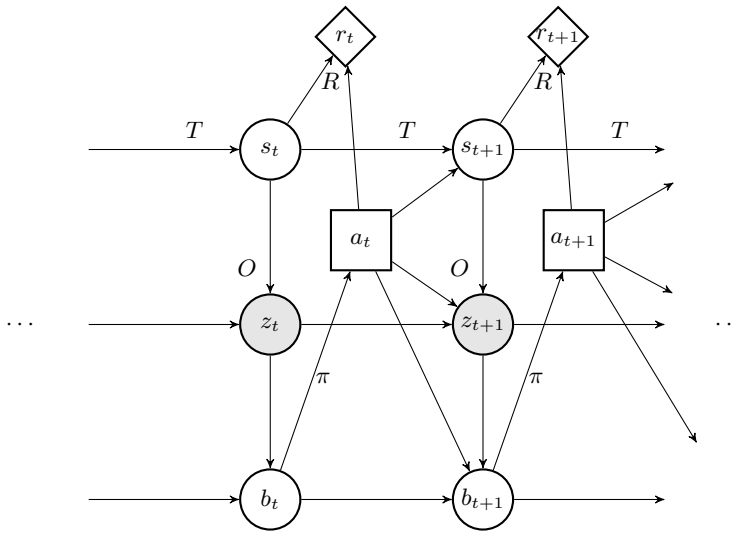

The influence diagram reflecting the graphical representation of the described railway maintenance problem is shown in Figure \ref{fig:pomdp_railway}. Compared to Figure \ref{fig:pomdp}, this graphical model presents arrows between observation variables displaying the autoregressive dependency. Autoregressive hidden Markov models (ARHMMs) have been deeply studied in the literature \cite{mor2021systematic} with application that range from the modelling of wind time-series \cite{ailliot2012markov}, to fault detection and prognostics tasks \cite{juesas2021autoregressive}. Similarly to our case, both works exploited ARHMMs to capture the switching between different internal states, while ensuring temporal coherence on observations stemming from sensor measurements. This is crucial in our case given the continuous nature of the derived fractal value measurements. This extension emphasizes the high flexibility of probabilistic graphical models, when incorporated into the POMDP schema. Moreover, the continuous dimension of the observations (fractal values) and the dependency among observations render this problem non-trivial to solve by means of common POMDP solution algorithms, which commonly assume $Z$ to be discrete. 
 
For the POMDP problem to be fully specified, i) the transition dynamics $T$ describing the probability $p(s_{t+1}|s_t,a_t)$ and ii) the observation model $O$ describing the likelihood $p(z_t|s_t,a_{t-1}, z_{t-1})$ must still be defined. 
We learn both models on the basis of the collected data of the fractal values indicator. The next section presents the employed methods and the inference results.

\section{Model inference}
\label{sec:inference}
The inference of the parameters governing the transition dynamics and the observation generating process serves to simulate and eventually optimize the railway maintenance planning problem. The recovered posterior distributions are conditioned on data presented in Section \ref{sec:data}, which comprise information on the data-derived fractal values time-series and the recorded maintenance actions over the tracks. 

The model used for the inference is a HMM conditioned on actions. The transition model is defined as follows:
\begin{align}
\begin{split}
    T_0 & \sim \dirichlet(\alpha_0)\\
    s_0 & \sim \categorical(T_0)\\
    T & \sim \dirichlet(\alpha_T)\\
    s_t | s_{t-1}, a_{t-1} & \sim \categorical(T)\\
\end{split}
\end{align}
where $T_0$ represents the initial probability state distributions, while $\alpha_0$ and $\alpha_T$ are the prior concentration parameters. $T_0$ is assigned a uniform flat prior, whereas $T$ is given a strongly informative prior to regularize the deterioration or the repairing process. In a transition matrix, the diagonal represents the probability to remain in the same state, while upper-right and lower-left triangles are associated with the probabilities of the system to deteriorate and improve its condition, respectively. As such, the transition matrix related to the action do-nothing, which describes the deterioration process of the system, is regularized with higher prior probabilities on the diagonal, lower on the upper-right triangle and near-zero on the lower-left triangle. In contrast, the transition matrices associated with maintenance actions present higher prior probabilities on the left triangle and near-zero on the right triangle - in order to inform the model that improvements of the system should follow a repair action - but no assumption on the magnitude of improvement.

The observation generating process differs on the basis of the assumed previous action, which can be either $a_0$ (deterioration process) or one of two possible maintenance actions $a_1$, $a_2$. The deterioration process is reflected in the observation model as a \textit{Truncated Student's t} process as follows:
\begin{align}\label{eq:det_process}
\begin{split}
    z_t - z_{t-1} & \sim \TS(\mu_{d|s_t}, \sigma_{d|s_t}, \nu_{d|s_t}, \ub=- z_{t-1})\\
\end{split}
\end{align}
The Student's t distribution assigns higher probabilities to tail events than, e.g., the Normal distribution. With a Gaussian likelihood, outliers would induce large shifts in the learned model, in an attempt to render tail events more likely. The Student's t distribution is thus here adopted to enhance robustness of the HMM inference to outliers, which are expected in real-world measurements. Nevertheless, the inference is still free to estimate a high value of degrees of freedom $\nu_{d|s_t}$ if the ``fat tail'' hypothesis is not correct. The difference among subsequent observations depends on the parameters $\mu_{d|s_t}$, $\sigma_{d|s_t}$, and $\nu_{d|s_t}$ which are state-dependent. An inferred negative value of $\mu_{d|s_t}$ will reflect the negative trend observed in the actual data, but the deterioration process is not forced to monotonically decrease, such that measurement errors are permissible. This implies that an observation can assume a value that is higher to a previous one even when no maintenance actions are taken. The distributions are truncated in $-z_{t-1}$, imposing the negative property of fractal values. The process in Equation \ref{eq:det_process} can be seen as a random walk with Truncated Student's t steps or as a particular case of an autoregressive process, where the autoregressive parameter is not learned \cite{knight1989limit}. In existing literature, the deterioration process is also often modeled as a Gamma process \cite{papakonstantinou2014planning}. This alternative approach has been tested herein, but led to common inference issues, such as divergence and non-identifiability. Consequently, we adopted a truncated Student's t process, which yielded improved inference results.

The repair process is correspondingly modeled as an autoregressive process with a truncated Student's t likelihood, so that, once again, only negative values are permissible:
\begin{align}
\begin{split}
    z_t  & \sim \TS(k_{r|a_{t-1}}*z_{t-1}+\mu_{r|s_t}, \sigma_{r|s_t}, \nu_{r|s_t}, \ub=0)\\
\end{split}
\end{align}
Specifically, the average improvement in fractal values of the repair process is controlled by an autoregressive action-dependent parameter $k_{r|a_t}$ and a state-dependent parameter $\mu_{r|s_t}$, with standard deviation $\sigma_{r|s_t}$. It is worth clarifying that if the repair process presents no autoregressive property, the model inference will simply assign values close to 0 to the parameter $k_{r|a_t}$.

Since we cannot know whether the first observation stems from a deterioration or a repair process, similarly to the inference of the first hidden state, we model it separately as follows:
\begin{align}
\begin{split}
    z_0  & \sim \TS(\mu_{s_{t_0}}, \sigma_{s_{t_0}}, \nu_{s_{t_0}}, \ub=0)\\
\end{split}
\end{align}

Finally, the aforementioned parameters that influence the observation generating process are defined as follows:
\begin{align}\label{eq:obs_parameters}
\begin{split}
    \mu_{d|s_t} & \sim \N(\Bar{\mu}_{\mu_{d|s_t}}, \Bar{\sigma}_{\mu_{d|s_t}})\\
    \sigma_{d|s_t} & \sim \TN(\Bar{\mu}_{\sigma_{d|s_t}}, \Bar{\sigma}_{\sigma_{d|s_t}}, \lb=0)\\
    \nu_{d|s_t} & \sim \G(\Bar{\alpha}_{\nu_{d|s_t}}, \Bar{\beta}_{\nu_{d|s_t}})\\
    \mu_{r|s_t} & \sim \TN(\Bar{\mu}_{\mu_{r|s_t}}, \Bar{\sigma}_{\mu_{r|s_t}}, \ub=0)\\
    \sigma_{r|s_t} & \sim \TN(\Bar{\mu}_{\sigma_{r|s_t}}, \Bar{\sigma}_{\sigma_{r|s_t}}, \lb=0)\\
    \nu_{r|s_t} & \sim \G(\Bar{\alpha}_{\nu_{r|s_t}}, \Bar{\beta}_{\nu_{r|s_t}})\\
    \mu_{s_{t_0}} & \sim \TN(\Bar{\mu}_{\mu_{s_{t_0}}}, \Bar{\sigma}_{\mu_{s_{t_0}}}, \ub=0)\\
    \sigma_{s_{t_0}} & \sim \TN(\Bar{\mu}_{\sigma_{s_{t_0}}}, \Bar{\sigma}_{\sigma_{s_{t_0}}}, \lb=0)\\
    \nu_{s_{t_0}} & \sim \G(\Bar{\alpha}_{\nu_{s_{t_0}}}, \Bar{\beta}_{\nu_{s_{t_0}}})\\
    k_{r|a_t} & \sim \betad(\Bar{\alpha}_{k_{r|a_t}}, \Bar{\beta}_{k_{r|a_t}})\\
\end{split}
\end{align}

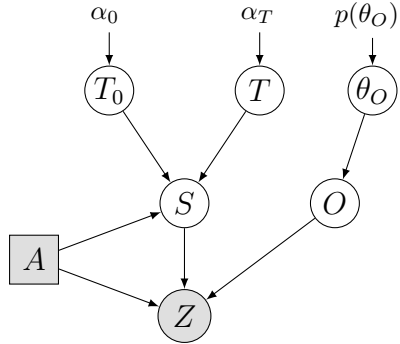
\begin{figure}[htb]
\centering
\begin{tikzpicture}
    \node [obs] (Z) at (0,0) {\large $Z$};
    \node [circle,draw=black,fill=white,inner sep=0pt,minimum size=0.65cm] (s) at (0,1.5) {\large $S$};
    \node [circle,draw=black,fill=white,inner sep=0pt,minimum size=0.65cm] (T0) at (-1,3.0) {\large $T_0$};
    \node [circle,draw=black,fill=white,inner sep=0pt,minimum size=0.65cm] (T) at (1,3.0) {\large $T$};
    \node [circle,draw=black,fill=white,inner sep=0pt,minimum size=0.65cm] (theta) at (2.5,3.0) {\large $\theta_O$};
    \node [circle,draw=black,fill=white,inner sep=0pt,minimum size=0.65cm] (O) at (2,1.5) {\large $O$};
    \node [rectangle,draw=black,fill=gray!25,inner sep=0pt,minimum size=0.65cm] (A) at (-2,0.75) {\large $A$};

    \node [text width=0.5cm] (a0) at (-1,4) {$\alpha_0$};
    \node [text width=0.5cm] (aT) at (1,4) {$\alpha_T$};
    \node [text width=1cm] (a) at (2.5,4) {$p(\theta_O)$};

    \path [draw,->] (T0) edge (s);
    \path [draw,->] (T) edge (s);

    \path [draw,->] (a0) edge (T0);
    \path [draw,->] (aT) edge (T);
    \path [draw,->] (theta) edge (O);
    \path [draw,->] (O) edge (Z);
    \path [draw,->] (s) edge (Z);
    \path [draw,->] (a) edge (theta);
    \path [draw,->] (A) edge (Z);
    \path [draw,->] (A) edge (s);


\end{tikzpicture}
\caption{Graphical model of the inferred HMM. For simplicity, we defined $\theta_O$ and $p(\theta_O)$ as the parameters of the observation model in Equation \ref{eq:obs_parameters} and their priors, respectively. Arrows indicate dependencies, while shaded nodes indicate observed variables.}
\label{fig:hmm}
\end{figure}

\begin{figure}[!hb]
    \centering
    \includegraphics[width=\linewidth]{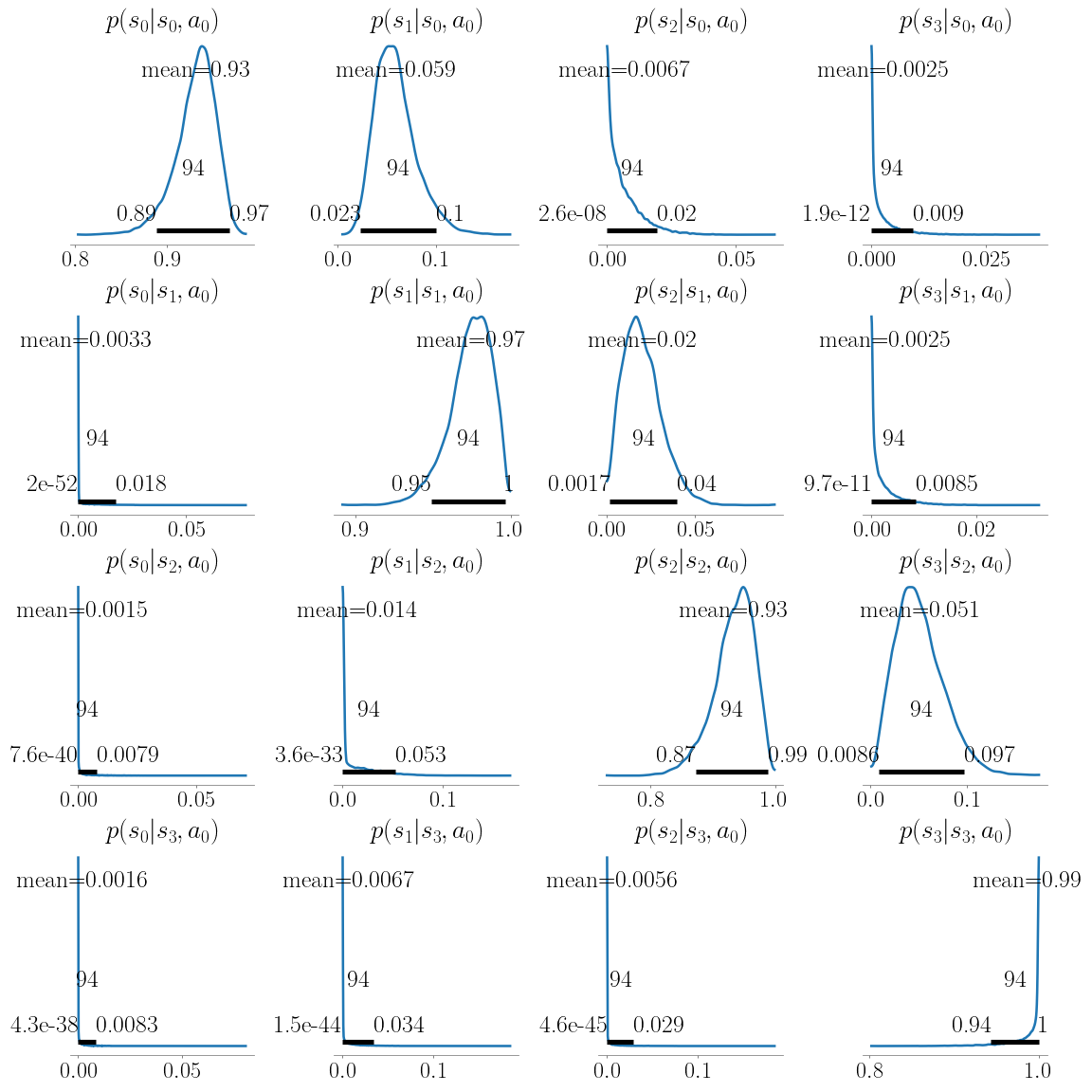}
    \caption{Transition matrix related to action do-nothing $a_0$. The distribution at row $i$ and column $j$ is associated with the probability to transition from state $i$ to $j$ when action $a_0$ is taken. Consistent with what is naturally expected in deterioration processes the highest probabilities are assigned to the state remaining invariant (diagonal entries), lower probabilities exist for deterioration transitions (upper right triangle), and almost zero probability is assigned to improvements of the system (lower left triangle).}
    \label{fig:tr_mat0}
\end{figure}

The entire HMM graphical model is displayed in Figure \ref{fig:hmm}, where shaded nodes indicate the observed variable provided from inspection data. Hidden variables are inferred by means of MCMC sampling exploiting a Hamiltonian Monte Carlo algorithm; namely the No-U-Turn sampler (NUTS) \cite{hoffman2014no}. The model is fed with the aforementioned fractal values and the performed actions. As stated in Section \ref{sec:data}, we have access to 10 years of recordings of fractal values and maintenance actions over several tracks. Fractal values are sampled twice per year every 2.5 meters, but maintenance actions produce effects over a much broader portion of the track. In addition, fractal values of such a small section are noisy. In order to mitigate these effects, we average fractal values every 150 meters. We finally build a dataset of 62 time-series, each one composed by 20 fractal values and 20 maintenance actions (action do-nothing included). As a result, one time-step of the POMDP problem is equal to 6 months. The inference is run with 4 chains and 3,000 samples collected after 4,000 burn-in samples per chain. The recovered posterior distributions present good post-inference diagnostic statistics, with no divergences and high homogeneity between and within chains.

\subsection{Inference results}

\begin{figure}[!h]
    \centering
    \includegraphics[width=\linewidth]{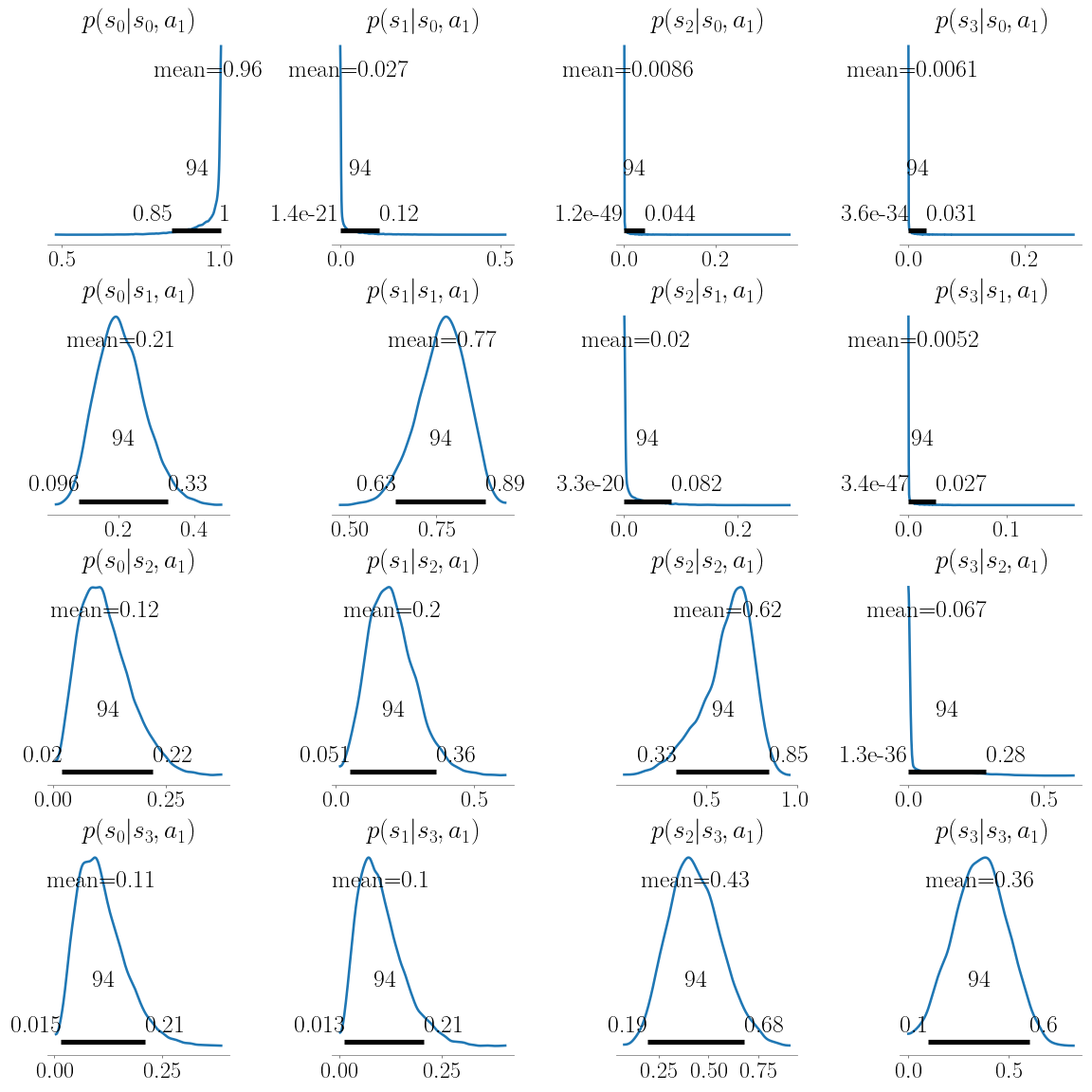}
    \caption{Transition matrix related to action $a_1$ (tamping). The distribution at row $i$ and column $j$ is associated with the probability to transition from state $i$ to $j$ when action $a_1$ is taken. Deterioration of the system (upper right traingle) reflects an almost zero probability, while it appears most
    probable to remain in the same condition or improve by a maximum of one state, which reflects the reduced influence of this action.}
    \label{fig:tr_mat1}
\end{figure}

The inferred transition matrix related to the action do-nothing $a_0$ is reported in Figure \ref{fig:tr_mat0}. Differently from the transition matrices shown in \cite{papakonstantinou2018pomdp}, for example, each entry is not a single parameter but a distribution of plausible values as a consequence of the robust formulation here and the MCMC inference. As seen in Figure \ref{fig:tr_mat0}, consistent with what is naturally expected in deterioration processes, the highest probability is assigned to remaining in the same state after one time-step (diagonal entries). A deterioration to the subsequent condition level is the second most likely transition, while improvements have near zero probability. Once the structure has reached the worst possible state, i.e., $s_3$, it stays in this condition with a probability that almost equals 1.

\begin{figure}[!h]
    \centering
    \includegraphics[width=\linewidth]{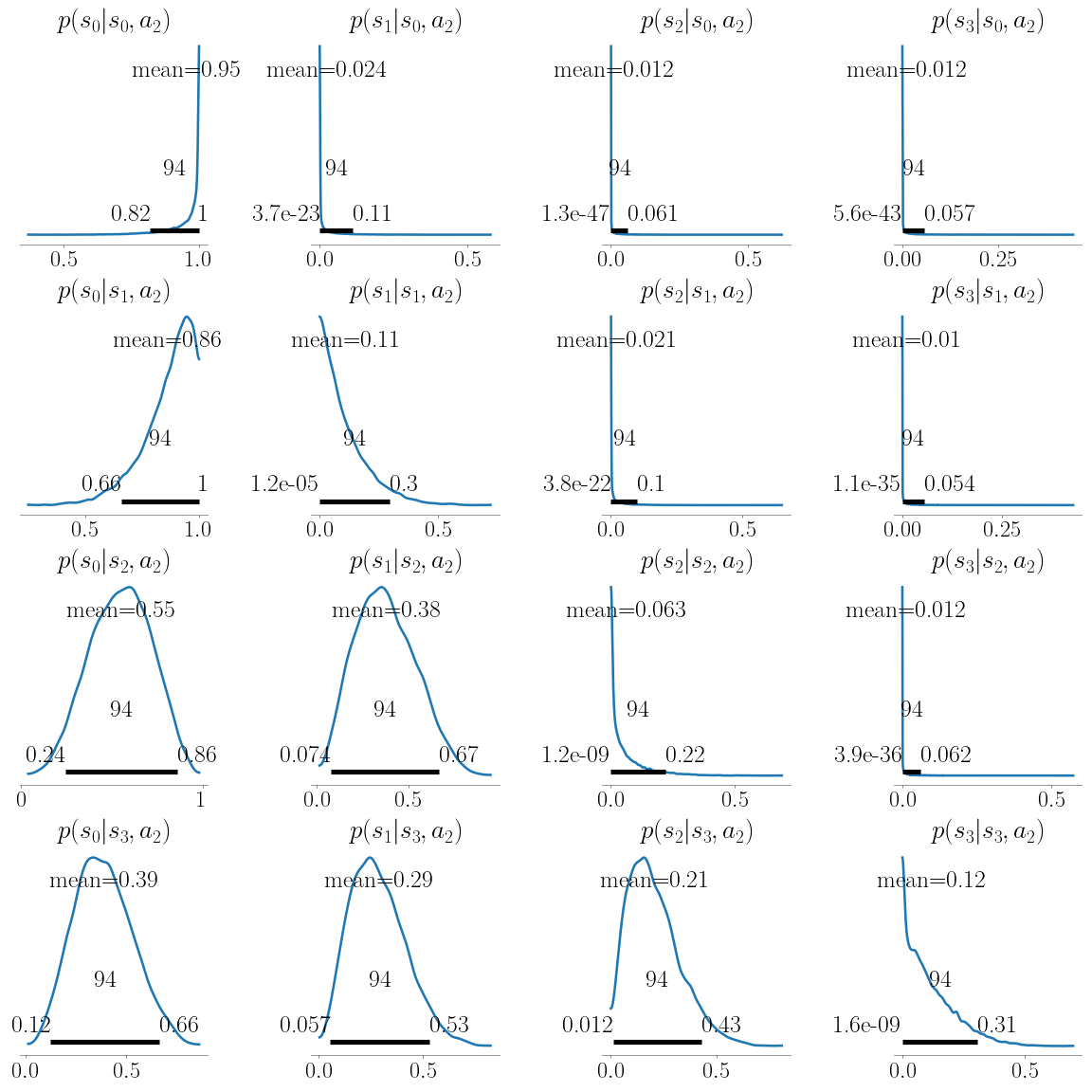}
    \caption{Transition matrix related to action $a_2$ (renewal plus tamping). The distribution at row $i$ and column $j$ is associated with the probability to transition from state $i$ to $j$ when action $a_2$ is taken. Transition to the best possible state $s_0$ is consistently assigned the highest probability, regardless of the starting state, reflecting the higher repairing effect of this maintenance action.}
    \label{fig:tr_mat2}
\end{figure}

Figure \ref{fig:tr_mat1} displays the transition matrix associated with the tamping action $a_1$, which is a low cost maintenance action with limited effect. If this action is assumed at state $s_0$, the environment stays in this condition with a probability almost equal to one (high certainty). For deteriorated states, it appears most probable to remain in the same condition or improve by a maximum of one state, although some smaller probabilities are assigned for larger improvements from state $s_2$ and $s_3$, which reflects the reduced influence of this action. Deterioration from any given state, upon assumption of such an action, reflects an almost zero probability. 

Figure \ref{fig:tr_mat2} displays the transition matrix associated with action $a_2$. Differently from the previous action, transition to the best possible state $s_0$ is consistently assigned the highest probability, regardless of the starting state. While we provided informative priors to regularize the deterioration or the repairing process, we stress that the MCMC inference learned this higher repairing effect of this maintenance action purely from data. It is worth mentioning that a lower probability of remaining in the same deteriorated state does exist, albeit substantially smaller than for action $a_1$, reflecting a ``failure'' of maintenance actions, which was also observed in the training data.

Finally, the observation model parameters are reported in \ref{app:inference_observation} in Figures \ref{fig:params_d}-\ref{fig:params_i}. It is worth noting the inferred results for the autoregressive parameter $k_{r|a_t}$. The distribution related to action $a_1$ comprises significantly higher values than the distribution associated with action $a_2$, highlighting that the fractal values are allowed to improve more when the latter is applied. While the two parameters were given the same prior, the MCMC inference still learned the substantial different effect of the two maintenance actions. Interestingly, the posterior distributions of the degrees of freedom suggest that the observations are especially ``far'' from being normally distributed during deterioration.

\begin{figure*}[h!]
\begin{subfigure}[b]{.5\textwidth}
\centering
\begin{tikzpicture}[scale=0.9]
\begin{axis}[
    width=6.8cm,height=6cm,
    title={\footnotesize Real data},
    xlabel={Timestep},
    ylabel={Fractal values},
    label style={font=\footnotesize},
    tick label style={font=\footnotesize},
    ymin=-1.7, ymax=0.2,
    ytick={-0.2,-0.6, -1.,-1.4},
    legend pos=south west,
    ymajorgrids=true,
    grid style=dashed,
    legend style={nodes={scale=0.7, transform shape}},
    y tick label style={
        /pgf/number format/.cd,
        fixed,
        fixed zerofill,
        precision=2,
        /tikz/.cd
    }
]

\addplot[
    color=black,
    mark=square,
    ]
    coordinates {
    (0,-0.15528443583050844)(1,-0.22509912216666667)(2,-0.32927595300000007)(3,-0.4303201339344263)(4,-0.52641)(5,-0.6225366169491525)(6,-0.7373024186885248)(7,-0.8141795173333328)(8,-0.9283017265000002)(9,-1.0961115398333334)(10,-1.1357525516666662)(11,-1.30102252295082)(12,-1.4154626147540985)(13,-0.37713225133333345)(14,-0.36788021316666664)(15,-0.37108925400000004)(16,-0.4460695331666667)(17,-0.5099044701694916)(18,-0.5291229250819672)
    };
    
\addplot[
    color=red,
    mark=otimes,
    ]
    coordinates {
    (0,-0.2)(1,-0.6)(2,-0.6)(3,-0.6)(4,-0.6)(5,-0.6)(6,-0.6)(7,-0.6)(8,-0.6)(9,-0.6)(10,-0.6)(11,-0.6)(12,-0.6)(13,0.2)(14,-0.2)(15,-0.2)(16,-0.2)(17,-0.2)(18,-0.2)
    };

\addplot[thick, samples=100,red]coordinates {
    (12,-1.7)(12,0.2)
    };
    \legend{fractal values,hidden states,action $a_1$}
\end{axis}

\end{tikzpicture}

\end{subfigure}
\begin{subfigure}[b]{.5\textwidth}
\centering
\begin{tikzpicture}[scale=0.9]
\begin{axis}[
    width=6.8cm,height=6cm,
    title={\footnotesize Simulated data},
    xlabel={Timestep},
    ymin=-1.7, ymax=0.2,
    ytick={-0.2,-0.6, -1.,-1.4},
    yticklabels={,,},
    label style={font=\footnotesize},
    tick label style={font=\footnotesize},
    legend pos=south west,
    ymajorgrids=true,
    grid style=dashed,
    y tick label style={
        /pgf/number format/.cd,
        fixed,
        fixed zerofill,
        precision=2,
        /tikz/.cd
    }
]

\addplot[
    color=black,
    mark=square,
    ]
    coordinates {
    (0,-0.2514891219513712)(1,-0.3720421069722283)(2,-0.46128843867032765)(3,-0.648814894446031)(4,-0.6662982729797794)(5,-0.7132892368631941)(6,-0.7510642421843678)(7,-0.8136913373526552)(8,-0.9957021611664494)(9,-1.0957726999433857)(10,-1.1329429044614774)(11,-1.2687874042909917)(12,-1.3476412763652563)(13,-0.30197364202425747)(14,-0.3206419203179265)(15,-0.24034568143011442)(16,-0.3055609679280589)(17,-0.3566620196154873)(18,-0.3671652910551904)
    };
    
\addplot[thick, samples=100,red]coordinates {
    (12,-1.7)(12,0.2)
    };
\end{axis}

\begin{axis}[hide x axis,axis y line*=right,ymin=-1.7,ymax=0.2,
width=6.8cm,height=6cm,
       ytick=\empty,
       extra y ticks={0.2,-0.2,-0.6,-1},
       extra y tick labels={0,1,2,3},
       ylabel={Hidden states},
       label style={font=\footnotesize},
    tick label style={font=\footnotesize},
       legend pos=south west,
       ]
  \addplot[
    color=red,
    mark=otimes,
    ]
    coordinates {
    (0,-0.2)(1,-0.6)(2,-0.6)(3,-0.6)(4,-0.6)(5,-0.6)(6,-0.6)(7,-0.6)(8,-0.6)(9,-0.6)(10,-0.6)(11,-0.6)(12,-0.6)(13,0.2)(14,-0.2)(15,-0.2)(16,-0.2)(17,-0.2)(18,-0.2)
    };
\end{axis}
\end{tikzpicture}

\end{subfigure}

\caption{One indicative time series of fractal values sampled from real data (left) and simulated parameters (right). A maintenance action $a_1$ was taken at timestep 12 in both cases. The associated hidden states are reported in red circles.}
        \label{fig:samples_env_1}
\end{figure*}
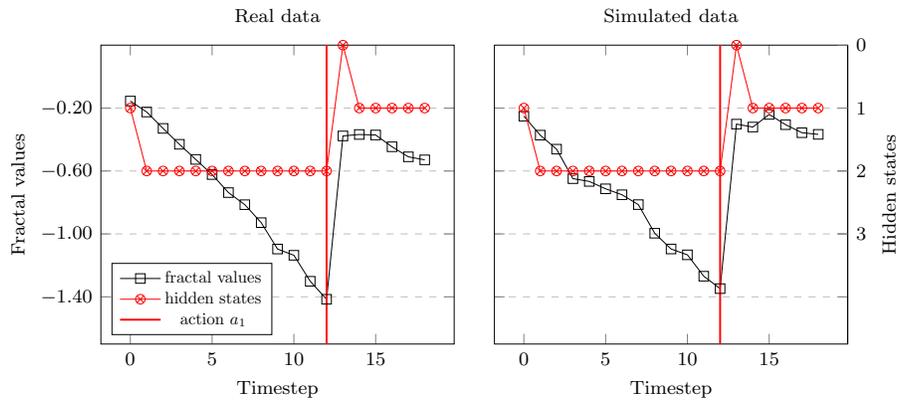

In order to further validate the goodness of the results, Figure \ref{fig:samples_env} compares an indicative time-series from real data with one sampled from inferred parameters, where starting values are close and no maintenance action was taken (pure deterioration process). Despite the stochasticity of the observations, the two time-series look extremely similar. Furthermore, it is possible to observe the slow variation of the underlying hidden states, as a result of the inferred transition matrix in Figure \ref{fig:tr_mat0}, which assigns the highest probability along the diagonal. Conditioning every observation on the previous value in the HMM allowed to correctly model the negative trend of the observations even in absence of changes in the hidden states. As a result, time-series of fractal values simulated from inferred parameters highly resemble the real data. A simpler purely non-autoregressive HMM would not have been able to capture this behavior and would have produced observations that would oscillate around some mean values. The inferred hidden state of the penultimate observation might be questionable and it is probably worth explaining. First, it should be noted that the trajectory of hidden states plotted is computed from the average across MCMC samples and jointly sampled during inference, i.e., each hidden state in the trajectory affects each other's inference. After a high number of deterioration time-steps, the likelihood of the state remaining invariant becomes significantly lower. When this likelihood becomes too low, but the state transition is not reflected in the observation, the inference might still assign the observation to the new state and explain the given value as an outlier/measurement error, which are indeed permissible thanks to the Student's t likelihood. This is exactly what occurred at the penultimate observation, for which the average MCMC samples revealed high uncertainty on whether the hidden state was $s_0$ or $s_1$. The inferred state transition is then further strengthened by the clearly visible fractal value jump, revealing the change to condition state $s_1$ in the following and final observation. Likewise, Figure \ref{fig:samples_env_1} shows real and simulated data, where a maintenance action was taken in similar conditions, in order to examine the goodness of the learned repair effect.

\begin{figure*}[ht]
\begin{subfigure}[b]{.5\textwidth}
\centering
\begin{tikzpicture}[scale=0.9]
\begin{axis}[
    width=6.8cm,height=6cm,
    title={\footnotesize Real data},
    xlabel={Timestep},
    ylabel={Fractal values},
    label style={font=\footnotesize},
    ymin=-0.5, ymax=0.,
    ytick={0,-0.1,-0.2,-0.3,-0.4, -0.5},
    tick label style={font=\footnotesize},
    legend pos=south west,
    ymajorgrids=true,
    grid style=dashed,
    legend style={nodes={scale=0.7, transform shape}}, 
    y tick label style={
        /pgf/number format/.cd,
        fixed,
        fixed zerofill,
        precision=2,
        /tikz/.cd
    }
]

\addplot[
    color=black,
    mark=square,
    ]
    coordinates {
    (0,-0.10805525625)(1,-0.12197251293220339)(2,-0.13966021466666667)(3,-0.14320736270000006)(4,-0.15605964672500006)(5,-0.16891193075000005)(6,-0.18407693483333334)(7,-0.18486231733333336)(8,-0.18628515983333335)(9,-0.22335532169491532)(10,-0.23092816472740108)(11,-0.2456097721666666)(12,-0.265463228)(13,-0.2543051906666666)(14,-0.25856303799999997)(15,-0.2811782294915253)(16,-0.2960109916666667)(17,-0.30161156866666666)(18,-0.3072121456666666)(19,-0.2931408418333333)(20,-0.3916938756666666)
    };

\addplot[
    color=red,
    mark=otimes,
    ]
    coordinates {
    (0,0)(1,0)(2,0)(3,0)(4,0)(5,0)(6,0)(7,0)(8,0)(9,0)(10,0)(11,0)(12,0)(13,0)(14,0)(15,0)(16,0)(17,0)(18,0)(19,-0.1)(20,-0.1)
    };
    \legend{fractal values, hidden states}
    
\end{axis}
\end{tikzpicture}

\end{subfigure}
\begin{subfigure}[b]{.5\textwidth}
\centering
\begin{tikzpicture}[scale=0.9]
\begin{axis}[
    width=6.8cm,height=6cm,
    title={\footnotesize Simulated data},
    xlabel={Timestep},
    ymin=-0.5, ymax=0.,
    ytick={0,-0.1,-0.2,-0.3,-0.4, -0.5},
    yticklabels={,,},
    label style={font=\footnotesize},
    tick label style={font=\footnotesize},
    legend pos=south west,
    ymajorgrids=true,
    grid style=dashed,
    y tick label style={
        /pgf/number format/.cd,
        fixed,
        fixed zerofill,
        precision=2,
        /tikz/.cd
    }
]

\addplot[
    color=black,
    mark=square,
    ]
    coordinates {
    (0,-0.13945465773216137)(1,-0.13832894651867395)(2,-0.14397455459280611)(3,-0.1540519717346652)(4,-0.16321266372315724)(5,-0.1749134728647964)(6,-0.18781890652227987)(7,-0.20133644652114352)(8,-0.20432108443428396)(9,-0.22599300266313704)(10,-0.23230371347235246)(11,-0.24071889308025188)(12,-0.23887299253613734)(13,-0.24924104409769446)(14,-0.25711537288815506)(15,-0.2723248134576546)(16,-0.28024079398979707)(17,-0.28404638226949885)(18,-0.2731024214657885)(19,-0.2799346512970818)(20,-0.36793888884741555)
    };

\end{axis}

\begin{axis}[hide x axis,axis y line*=right,ymin=-0.5,ymax=0,
width=6.8cm,height=6cm,
       ytick=\empty,
       extra y ticks={0,-0.1,-0.2,-0.3,-0.4, -0.5},
       extra y tick labels={0,1,2,3},
       ylabel={Hidden states},
       label style={font=\footnotesize},
       tick label style={font=\footnotesize},
       legend pos=south west,
       ]
  \addplot[
    color=red,
    mark=otimes,
    ]
    coordinates {
    (0,0)(1,0)(2,0)(3,0)(4,0)(5,0)(6,0)(7,0)(8,0)(9,0)(10,0)(11,0)(12,0)(13,0)(14,0)(15,0)(16,0)(17,0)(18,0)(19,-0.1)(20,-0.1)
    };
\end{axis}
\end{tikzpicture}

\end{subfigure}

\caption{One indicative time series of fractal values sampled from real data (left) and simulated parameters (right). No maintenance action was taken in the two samples. The associated hidden states are reported in red circles.}
        \label{fig:samples_env}
\end{figure*}
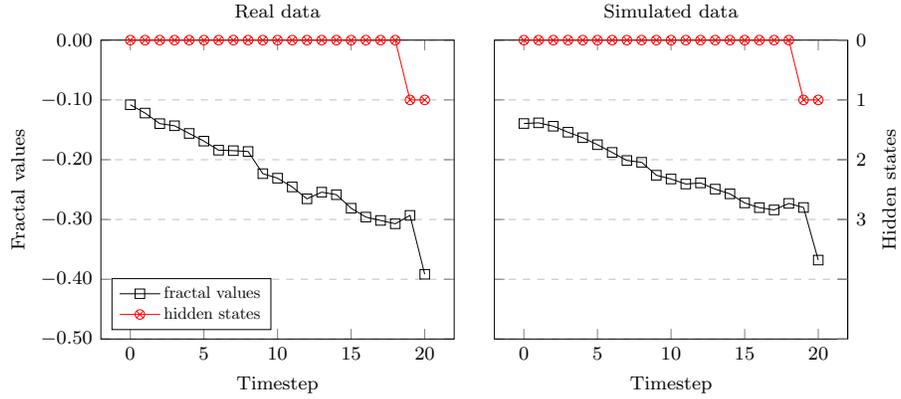

While we can not directly provide the code of the inference presented in this section to protect railway data provided by SBB, we provide a tutorial\footnote{Code available on \href{https://github.com/giarcieri/Hidden-Markov-Models/blob/1b6cf56d2407c9cc7a7315ae76b1896f23110d00/HMMs for deterioration process Truncated Normal Process.ipynb}{GitHub}.} on the inference run on simulated data that resembles our recordings. The tutorial shows how to recover transition and observation model for MDP and POMDP cases. We hope that such a practice will favor the modeling of MDP and POMDP settings based on data and will further support their utilization in  solutions for real-world applications.

Several extensions to the HMM used in this work are possible. For instance, a Bayesian hierarchical model \cite{gelman1995bayesian} could be applied to allow dependencies between components of the system \cite{memarzadeh2016hierarchical, luque2019risk, andriotis2019managing, morato2022inference}. In our case, we may model dependencies between closer tracks that may be affected by similar substructure deterioration. Moreover, one may extend the HMM to time-dependent transition matrices, if there is evidence that the parameters governing the dynamics change over time. Transition matrices would be then enlarged by a further dimension representing time, which would be encoded in the solution. Significant amount of additional data would, however, be required in such a case in order to adequately estimate the model parameters.

\section{Solving the POMDP}
\label{sec:solution}

After having inferred all model parameters, it is now possible to solve the optimization problem, namely to find the optimal policy to be executed given states or observations. 

\subsection{Full observability}
\label{sec:solution_full}

First, full observability of the problem is assumed, i.e., the optimal policy is computed for the case when states are directly and accurately observed. This allows to draw an upper bound of the performance that the POMDP solution can achieve. We consider an infinite horizon problem, with $\gamma=0.995$, and apply the Q-value iteration algorithm (Equation \ref{eq:q_value}) over the entire model distributions, represented by 12,000 samples, that in the MDP case here considered coincides with $p(T)$. By implementing the algorithm with JAX \cite{frostig2018compiling, jax2018github}, a JIT compiler for generating high-performance accelerator code, solving the problem for the entire inferred distributions takes only a handful of seconds. As a result, Q-value distributions are computed. Applying Equation \ref{eq:bayes_optimal_action_q}, it is possible to take an expectation and compute the optimal action for each state. The resulting actions are thus optimal for the entire range of parameter distributions, i.e., they are robust over epistemic uncertainty. The optimal actions are reported in Table \ref{tab:optimal_action}. Interestingly, if one applies Q-value iteration algorithm to only the mean values of the inferred transition matrices, i.e., discarding all information contained in the posterior distributions, a different optimal policy is obtained. The policy optimized with the mean parameters estimates is also reported in Table \ref{tab:optimal_action}.

\begin{table}[!t]
\caption{Optimal action for every state, optimized for all posterior distributions (top) and only for the mean values (bottom).}
    \centering
    \begin{tabular}{lcccc}
        \multicolumn{1}{c}{\bf State condition}  &\multicolumn{1}{c}{\bf $s_0$}  &\multicolumn{1}{c}{\bf $s_1$} &\multicolumn{1}{c}{\bf $s_2$}
        &\multicolumn{1}{c}{\bf $s_3$}\\
        \hline
        \bf Robust optimal action & $a_0$ & $a_1$ & $a_1$  & $a_2$ \\
        \bf Optimal action with posterior mean & $a_0$ & $a_1$ & $a_2$  & $a_2$ \\
    \end{tabular}
    \label{tab:optimal_action}
\end{table}

\begin{figure*}[!ht]
\begin{subfigure}[b]{0.5\textwidth}
\centering
\begin{tikzpicture}[scale=0.9]
\begin{axis}[
    width=7cm,height=6cm,
    title={\footnotesize State $s_0$},
    xlabel={Action},
    ymin=0, ymax=12000,
    ytick={0,4000,8000,12000},
    label style={font=\footnotesize},
    tick label style={font=\footnotesize},
    ymajorgrids=true,
    grid style=dashed,
	x tick label style={
		/pgf/number format/1000 sep=},
	ylabel={\# optimal action},
	enlargelimits=0.05,
	ybar interval=0.7,
]
\addplot 
	coordinates {(2,0) (1,5677) (0,6323) (-1,0)};

\end{axis}
\node[below] at (1.1,2.85) {\footnotesize $53\%$};
\node[below] at (2.8,2.6) {\footnotesize $47\%$};
\node[below] at (4.5,0.8) {\footnotesize $0\%$};
\end{tikzpicture}
\end{subfigure}%
\begin{subfigure}[b]{0.5\textwidth}
\centering
\begin{tikzpicture}[scale=0.9]
\begin{axis}[
    width=7cm,height=6cm,
    title={\footnotesize State $s_1$},
    xlabel={Action},
    ymin=0, ymax=12000,
    ytick={0,4000,8000,12000},
    label style={font=\footnotesize},
    tick label style={font=\footnotesize},
    ymajorgrids=true,
    grid style=dashed,
	x tick label style={
		/pgf/number format/1000 sep=},
	enlargelimits=0.05,
	ybar interval=0.7,
]
\addplot 
	coordinates {(2,195) (1,11008) (0,797) (-1,0)};

\end{axis}
\node[below] at (1.1,1.) {\footnotesize $7\%$};
\node[below] at (2.8,4.4) {\footnotesize $92\%$};
\node[below] at (4.5,0.85) {\footnotesize $0.2\%$};
\end{tikzpicture}
\end{subfigure}%
\hfill
\begin{subfigure}[b]{0.5\textwidth}
\centering
\begin{tikzpicture}[scale=0.9]
\begin{axis}[
    width=7cm,height=6cm,
    title={\footnotesize State $s_2$},
    xlabel={Action},
    ymin=0, ymax=12000,
    ytick={0,4000,8000,12000},
    label style={font=\footnotesize},
    tick label style={font=\footnotesize},
    ymajorgrids=true,
    grid style=dashed,
	x tick label style={
		/pgf/number format/1000 sep=},
	ylabel={\# optimal action},
	enlargelimits=0.05,
	ybar interval=0.7,
]
\addplot 
	coordinates {(2,4475) (1,7525) (0,0) (-1,0)};

\end{axis}
\node[below] at (1.1,0.8) {\footnotesize $0\%$};
\node[below] at (2.8,3.25) {\footnotesize $63\%$};
\node[below] at (4.5,2.2) {\footnotesize $37\%$};
\end{tikzpicture}
\end{subfigure}%
\begin{subfigure}[b]{0.5\textwidth}
\centering
\begin{tikzpicture}[scale=0.9]
\begin{axis}[
    width=7cm,height=6cm,
    title={\footnotesize State $s_3$},
    xlabel={Action},
    ymin=0, ymax=12000,
    ytick={0,4000,8000,12000},
    label style={font=\footnotesize},
    tick label style={font=\footnotesize},
    ymajorgrids=true,
    grid style=dashed,
	x tick label style={
		/pgf/number format/1000 sep=},
	enlargelimits=0.05,
	ybar interval=0.7,
]
\addplot 
	coordinates {(2,6825) (1,5175) (0,0) (-1,0)};

\end{axis}
\node[below] at (1.1,0.8) {\footnotesize $0\%$};
\node[below] at (2.8,2.45) {\footnotesize $43\%$};
\node[below] at (4.5,2.98) {\footnotesize $57\%$};
\end{tikzpicture}
\end{subfigure}%
\caption{Number of model samples for which each action is optimal at a given state. It allows to consider how optimal each action is with respect to the model distributions.}
\label{fig:optimal_action_counts_tikz}
\end{figure*}
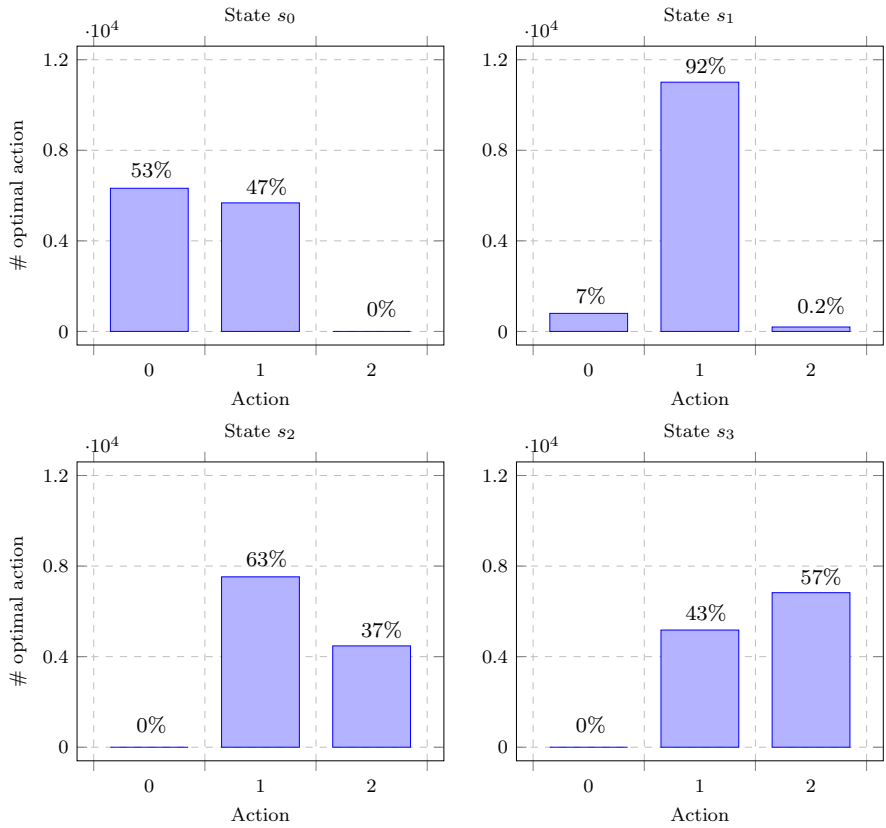

Moreover, by considering the full transition dynamics distribution, it is possible to compute the percentage of samples for which a specific action is optimal. The results are displayed in Figure \ref{fig:optimal_action_counts_tikz}. This allows to consider how confident one can be about action optimality, and highlights that it is very likely to obtain a different optimal policy if one optimizes for only a single sample of the transition model distribution. For instance, action $a_1$ is still optimal in $47\%$ and $43\%$ of samples when the system is in state $s_0$ and $s_3$, respectively. As a result, a policy where a tamping action is taken at every decision step is optimal for a significant number of samples of the inferred distributions. Exploiting the whole distribution parameter space turns out to be crucial for enhancing robustness of the computed policy.

Finally, we simulate the problem 20,000 times with 50 time-steps for all transition model samples, for a total of 240 millions possible trials. The robust optimal policy shown in Table \ref{tab:optimal_action} is then applied over all simulations. As a comparison, we also show results for the policy optimized only over the mean values and for the policy that always chooses action $a_1$. Results are reported in Table \ref{tab:results_mdp} in terms of average costs, Standard Error (SE), and $95\%$ Highest Density Interval (HDI). The robust optimal policy delivers the best expected result, although we clarified in Section \ref{sec:bayesian_decision_making} that it does not necessarily have to be the best one for any specific value of the model parameters.

\begin{table}[ht]
\caption{Expected total life-cycle costs of the robust optimal policy, the policy optimal for the mean values of transition matrices distributions, and a policy which chooses always action $a_1$ over 240 millions simulations.}
    \centering
    \begin{tabular}{lcccc}
        & {\bf Average} & {\bf SE} & {\bf HDI $2.5\%$} & {\bf HDI $97.5\%$} 
        \\
        \hline
        {\bf Robust optimal policy} & -13,377 & 0.67 & -33,700 & -5,000 
        \\
        {\bf Optimal action with posterior mean} & -13,493 & 0.60 & -31,600 & -5,000 
        \\
        {\bf Policy always $a_1$} & -16,072 & 0.94 & -46,300 & -7,500
    \end{tabular}
    \label{tab:results_mdp}
\end{table}

\subsubsection{Finite horizon}

Concluding the MDP solution study, we compute and showcase the optimal policy considering a finite horizon problem of $H=50$ time-steps, with terminal value of 0. The Q-value iteration algorithm applied over all inferred distribution parameters now computes $S\times A\times H$ distributions. Similarly to the infinite case, the optimal action at each time-step $t$ is computed as follows:

\begin{equation}\label{eq:bayes_optimal_action_q_t}
    a^*_t=\argmax_{a\in A}\E_{\mathbf{\theta}\sim p(\mathbf{\theta})}\left[Q^{\pi^*}_{\mathbf{\theta}}(s, a, t)\right]
\end{equation}

\begin{figure}[htb]
\begin{tikzpicture}
  \node (img)  {\includegraphics[width=0.95\textwidth]{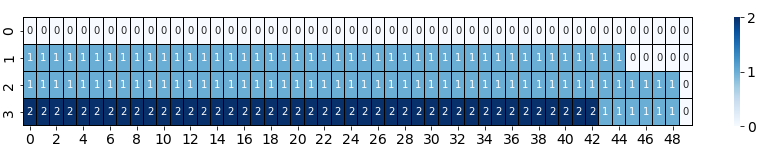}};
  \node[right=of img, node distance=0cm, rotate=270, anchor=center,xshift=-0.0cm, yshift=-1.cm] {\footnotesize Action};
  \node[left=of img, node distance=0cm, rotate=90, anchor=center,yshift=-0.9cm] {\footnotesize State};
 \end{tikzpicture}

\begin{tikzpicture}
  \node (img)  {\includegraphics[width=0.95\textwidth]{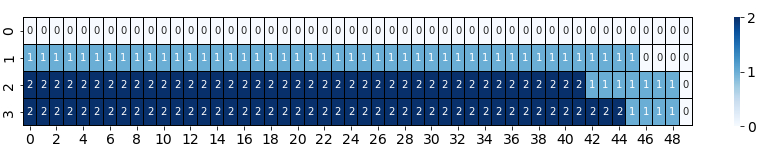}};
  \node[right=of img, node distance=0cm, rotate=270, anchor=center,yshift=-1.cm] {\footnotesize Action};
  \node[below=of img, node distance=0cm, yshift=1cm] {\footnotesize Timestep};
  \node[left=of img, node distance=0cm, rotate=90, anchor=center,yshift=-0.9cm] {\footnotesize State};
\end{tikzpicture}
\caption{Optimal policy considering all model distribution (above) and only the mean parameters (bottom) for any given state in time for a finite horizon of 50 time-steps.}
\label{fig:finite_horizon2}
\end{figure}
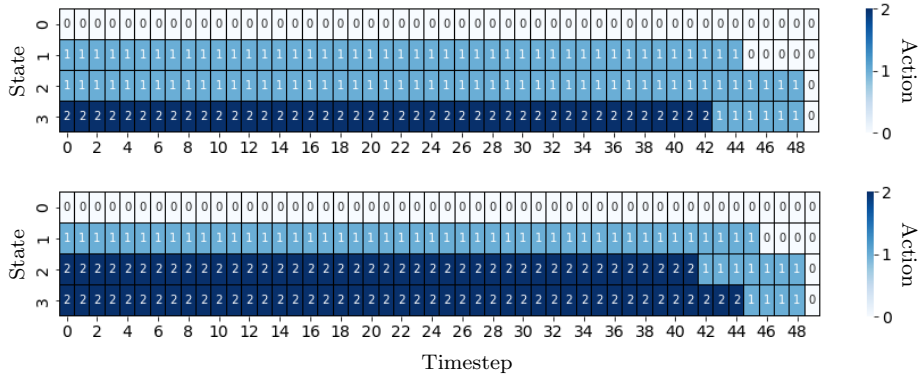

The resulting policy is reported in Figure \ref{fig:finite_horizon2}. Consistently with the infinite horizon case, solving the Bellman equation for the mean values of the inferred distributions leads to different results, especially for state $s_2$, further highlighting the importance of incorporating epistemic uncertainty into the solution.

\subsection{Partial observability}
This section presents now the solution to the POMDP problem. The states are hidden variables and the agent forms a belief over states given the observations received. As pointed out in Section \ref{sec:bg_pomdp}, planning an optimal policy through beliefs is a far more challenging task and point-based value iteration algorithms offer approximate solutions. However, solving POMDP problems with continuous observations remains an even more challenging task, even for these methods, which rely on discretization of the observation space. While some recent advances have been achieved to extend POMDP solvers to continuous observations, e.g. in \cite{hoerger2020line},
in this work we rely on a simpler applicable method called $Q_{MDP}$ \cite{cassandra1996acting}:

\begin{equation}\label{eq:qmdp}
    \pi_{Q_{MDP}}
    =\argmax_{a\in A}\sum_{s\in S}b(s)Q^{\pi^*}(s, a)
\end{equation}

Namely, the $Q_{MDP}$ method ignores the observation model and computes the Q-values of the underlying MDP given the transition model. It then finds the optimal action at each step by only updating the belief $b(s)$ with Equation \ref{eq:belief}. This results in extremely low computational load when compared to point-based methods, at the expense of reduced accuracy, in general problems.

Extending this algorithm to all inferred distributions to account for epistemic uncertainty is then straightforward:

\begin{equation}\label{eq:qmdp_uncertain}
    a^*=\argmax_{a\in A}\E_{\mathbf{\theta}\sim p(\mathbf{\theta})}\left[\sum_{s\in S}b_{\mathbf{\theta}}(s)Q^{\pi^*}_{\mathbf{\theta}}(s, a)\right]
\end{equation}

\noindent where both the Q-values and the beliefs depend on $\mathbf{\theta}$, which is a sample of the entire POMDP model from transition and observation parameter distributions. All computations among different samples are independent and thus easily parallelizable, without substantially increasing the computational load.

The $Q_{MDP}$ method assumes that the agent's observation uncertainty is removed after one step, in which case the method would provide the optimal solution. Thus, the agent always chooses the action associated with the highest long-term reward, for the current level of uncertainty. Based on this assumption, the main drawback of the method is that it does not choose information gathering actions. In cases where the POMDP problem comprises these actions, the transition dynamics are fast, and/or the observation uncertainty is significant, the method may result in poor performance \cite{papakonstantinou2014planning2}, otherwise it might be remarkably effective in some settings \cite{littman1995learning}. The reasons why the $Q_{MDP}$ method is especially suited to the studied problem in this work and further insights on the quality of the solution are drawn in Section \ref{sec:qmdp_quality}.

\subsubsection{Numerical results}

As in Section \ref{sec:solution_full}, we simulate several trials from the POMDP parameter distributions to obtain the average performance under nearly all possible scenarios over a finite horizon of 50 time-steps. At the beginning of every simulation, a different POMDP configuration is sampled from the parameter distributions and kept fixed over the 50 time-step horizon. However, the agent does not access the sampled transition and observation model parameters to compute the optimal policy, but it exploits all inferred parameter distributions, approximated through samples. The agent thus solves 12,000 POMDP problems in parallel, i.e., it computes distributions of solutions, and selects actions that maximize the expected value with respect to the entire model parameter
distribution, according to Equation \ref{eq:qmdp_uncertain}. The resulting policy is thus robust to parameter uncertainty and it does not need to access the actual POMDP environment parameters. This scheme aims to resemble a real-world scenario, where the agent would never access the real-world true parameters, and it needs to tackle this additional uncertainty. We show that, instead of assuming a particular sample of the POMDP parameter distributions as ground truth, our robust policy represents a natural and safer choice against model uncertainty. We compare the robust policy with the policy based on the means of the posterior parameter distributions, as well as five other different agents that assume knowledge of specific POMDP samples. Specifically, we order our samples based on their (unnormalized) posterior probability and select the ones that correspond to the $\left[0,25,50,75,100\right]$ percentiles. Each of the five agents computes the optimal policy with respect to the sample of the associated percentile. The resulting policies are then evaluated with the scheme previously described. Table \ref{tab:results_pomdp} summarizes the results in terms of mean performance, SE, and $95\%$ HDI over 100k simulations. The table also reports the results of the policy that always chooses actions $a_1$, already shown in Table \ref{tab:results_mdp}. It is worth noting that the standard error of the mean performance of this latter policy is only lower due to the larger number of evaluated simulations, as reported in Section \ref{sec:solution_full}. The robust policy achieves better mean performance than the other five benchmarking solutions based on specific samples. In particular, among the latter, three benchmarking policies prove not substantially better than the policy that always chooses action $a_1$. Even though the results from the policies associated with the percentiles 0 and 50 are not very distant from the results of the robust policy, this can not be known a priori. Likewise, it should not be surprising that the policy based on the posterior mean parameters even shows slightly better empirical results than the robust policy. Indeed, it can be possible to find specific samples, among all possible values, that perform similarly or even slightly better than the robust policy. However, this is strongly dependent on the shape of the parameter distributions and the assumed cost matrix, and one can not know the performance of such samples until they are actually evaluated. As such, in the context of model uncertainty, the safest choice is represented by the robust policy, which is optimized over all POMDP parameter samples, namely it is ``robust'' from a model uncertainty perspective. 

\begin{table}[t]
\caption{Expected total life-cycle costs associated with the $Q_{MDP}$ robust policy, the policy based on posterior mean parameters, and other five benchmarking policies computed from specific samples of the POMDP parameter distributions, evaluated over a finite horizon of 50 time-steps and 100k simulations of the POMDP environment.}
    \centering
    \begin{tabular}{lcccc}
        & {\bf Average} & {\bf SE} & {\bf HDI $2.5\%$} & {\bf HDI $97.5\%$} 
        \\
        \hline
        {\bf Robust policy} & -14,526 & 39.47 & -39,750 & -5,050 
        \\
        {\bf Policy with posterior means} & -14,478 & 37.68 & -38,250 & -5,050
        \\
        {\bf Policy with percentile 0} & -14,590 & 39.66 & -40,000 & -5,050
        \\
        {\bf Policy with percentile 25} & -15,933 & 45.15 & -45,420 & -7,400
        \\
        {\bf Policy with percentile 50} & -14,912 & 41.68 & -40,350 & -5,100
        \\
        {\bf Policy with percentile 75} & -15,822 & 48.4 & -44,950 & -7,200
        \\
        {\bf Policy with percentile 100} & -16,497 & 50.51 & -47,010 & -7,100
        \\
        {\bf Policy always $a_1$} & -16,072 & 0.94 & -46,300 & -7,500
    \end{tabular}
    \label{tab:results_pomdp}
\end{table}

Despite the partial observability, the robust policy in this case also delivers only slightly worse performance than the full observability case. We note that the agent's belief converges to the actual hidden states within a handful of observations in this problem, after which the actions taken are nearly always optimal. At initiation of the horizon the agent exhibits a conservative behavior by mostly choosing action $a_1$, which is indeed the most likely to be optimal when the uncertainty over the state is high. As a result, the observed disparity in the MDP versus the POMDP performance is primarily due to early decisions, when the agent's belief is not yet accurate. 

In all time-steps, the agent persistently selects actions that are optimal considering all parameter distributions, i.e., robust to epistemic uncertainty. Such a policy is agnostic with respect to real environment transition dynamics and observation generating processes. As such, computed solutions do not overfit a specific POMDP model configuration and are more likely to perform well when deployed to real-world applications, where the environment remains uncertain.

\subsection{On the quality of the $Q_{MDP}$ solver}\label{sec:qmdp_quality}

While our focus is not shed on the type of solver to adopt for solution of the POMDP problem, different solution techniques can have an important impact on the accuracy of the POMDP results. We thus further explain here the reasons why the simple $Q_{MDP}$ solver is particularly suited to this specific case. This is tied to the following problem traits: i) the permanent monitoring nature of the problem, namely the agent does not have to take information-gathering actions, but informative observations are provided at every time-step, ii) the continuous dimension of the observations, that would negatively impact the performance of more sophisticated algorithms, which would rely on discretization of the observation space, while this observations attribute is compatible with and does not affect the performance of the $Q_{MDP}$ solver, and iii) the high probabilities on the diagonal of the transition matrix associated with action $a_0$ do-nothing (Figure \ref{fig:tr_mat0}), which result in slow variation of the underlying hidden states, allowing the $Q_{MDP}$ solver to accurately detect the true hidden states with a few measurements (often only one). In absence of any of those characteristics, the $Q_{MDP}$ solver would not have been equally effective. 

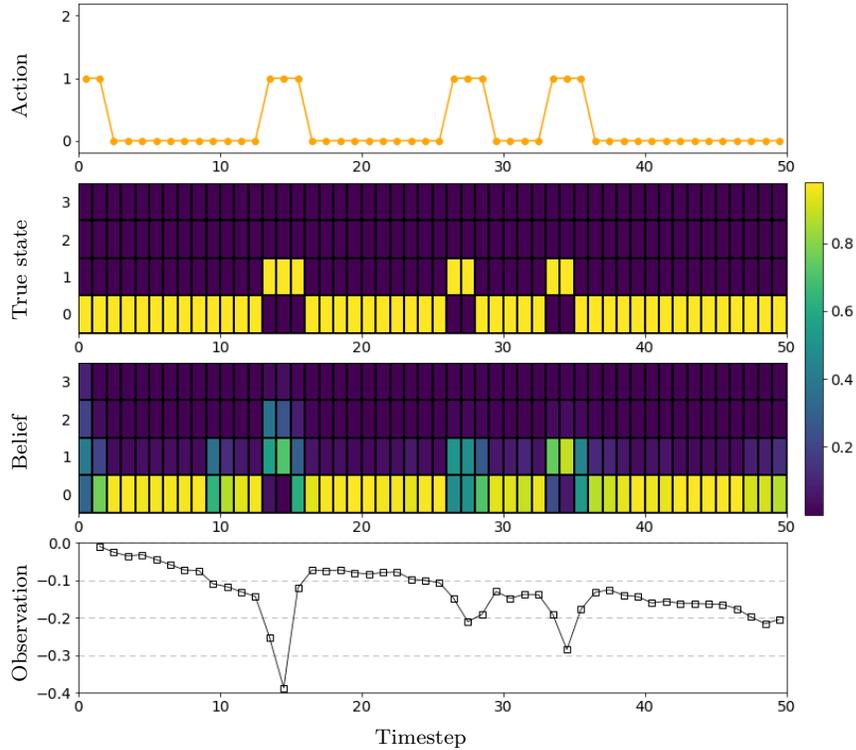
\begin{figure}[htb!]
\begin{tikzpicture}
  \node (img)  {\includegraphics[width=\textwidth]{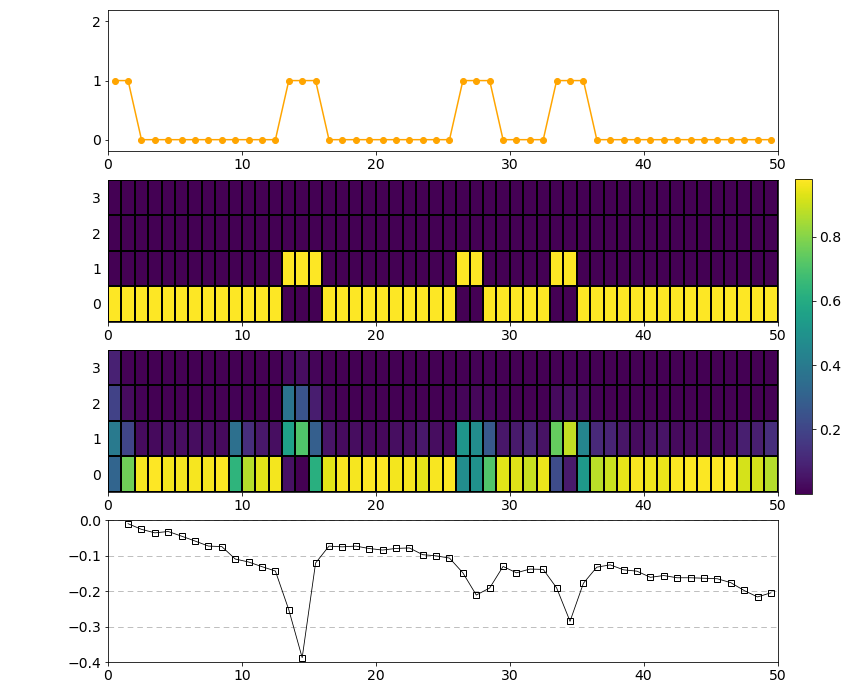}};
  \node[left=of img, node distance=0cm, rotate=90, anchor=center,xshift=3.6cm, yshift=-1.85cm] {\footnotesize Action};
  \node[below=of img, node distance=0cm,xshift=0cm,  yshift=1.1cm] {\footnotesize Timestep};
  \node[left=of img, node distance=0cm, rotate=90, anchor=center,xshift=1.15cm, yshift=-1.85cm] {\footnotesize True state};
  \node[left=of img, node distance=0cm, rotate=90, anchor=center,xshift=-1.15cm,yshift=-1.85cm] {\footnotesize Belief};
  \node[left=of img, node distance=0cm, rotate=90, anchor=center,xshift=-3.55cm,yshift=-1.85cm] {\footnotesize Observation};
 \end{tikzpicture}
\caption{A sample from the $Q_{MDP}$ planner. From bottom to top: the observations (fractal values) that the agent receives; the agent's beliefs based on the observations; the true hidden states (not accessed by the agent); the planned actions based on the agent's beliefs. After only one observation, the agent’s belief already detects the true hidden state and perfectly converges with the second observation. Despite the partial observability, the agent almost always plans the optimal action.}
\label{fig:qmdp_graph}
\end{figure}

In this section, we illustrate some practical findings to prove the quality of the solver for this POMDP problem, with a finite horizon setting. A key evidence is represented by the comparison of the theoretical value for the initial belief $V_{Q_{MDP}}(b_0)$ with the mean simulated reward achieved, where: 

\begin{equation}\label{eq:Vqmdp}
    V_{Q_{MDP}}(b_0)=\max_{a\in A}\sum_{s\in S}b_0(s)Q^{\pi^*}(s, a, t=0)
\end{equation}

\noindent The theoretical value is an optimistic upper bound, while the simulated value represents a lower bound (it is not the optimal policy). If the two are sufficiently close, that can be a good indication that  the $Q_{MDP}$ policy is close to the optimal one. In order to only evaluate the quality of the $Q_{MDP}$ planner, we fix the POMDP model parameters to their means. The theoretical value $V_{Q_{MDP}}(b_0)$ expects costs equal to -13,405 (upper bound), while 100k simulations of the $Q_{MDP}$ finite horizon policy over 50 time-steps, and optimized over the means of the model parameters, achieve an average cost of -14,374 (lower bound). Considering the high variability of the costs depending on the realized states (the simulations achieve -5,050 and -123,800 in the best and worst case scenario, respectively), the difference between the two bounds is quite tight.

As a further example, Figure \ref{fig:qmdp_graph} displays a sample trial of the $Q_{MDP}$ planner. The bottom figure shows the observations, i.e., the fractal values that the agent receives over the trial. Based on the observations, the agent forms beliefs over the states (third subplot), which are compared against the true hidden states, shown in the second subplot. Based on the beliefs, the agent plans the optimal actions, reported in the top subplot.  The belief is initialized according to the initial probability state distribution $T_0$ and is hence not accurate at the beginning. After only one observation, the agent's belief already largely detects the true hidden state and perfectly converges with the second observation. Afterwards, it remains extremely accurate. The agent is able to correctly detect the change to state $s_1$ at time-step 14 and plan the optimal action $a_1$ until the state returns to $s_0$. For other two consecutive times, the agent accurately and timely detects the deterioration to state $s_1$. In both cases, the state returns to perfect condition $s_0$ after 2 time-steps. However, the agent is uncertain about the correct state (whether it is $s_0$ or still $s_1$) and prefers to precautionary take a further third maintenance action $a_1$ to be sure of the improvement of the condition. These two decisions represent the only two instances, where after the initial warm up time post-initialization, the agent does not plan the optimal action under actual observation of the hidden state; this deficiency is owed to the uncertainty in the observations.


\section{Conclusions}
\label{sec:conclusion}

In this work, a maintenance planning problem is modeled and solved by means of a POMDP framework. A main contribution of this work is the demonstration of the end-to-end inference of the POMDP model purely from available data. We showcase our method on a real-world maintenance planning problem for railway track infrastructure. We exploit real-world observations (monitoring data) in the form of computed fractal values and actual maintenance actions recorded across Switzerland's railway network. We apply a hidden Markov model conditioned on actions, relying on a truncated Student's t process which describes the deteriorating system, to infer the transition dynamics and the observation generating process of the POMDP problem. Parameter distributions that represent all plausible values under the available data are inferred through MCMC sampling of the model, exploiting the NUTS algorithm. The results present high evidence of convergence, with the simulations highly resembling the real data.

A further contribution of this work lies in application of the inferred model parameter distributions for solving the maintenance planning problem, i.e., computing the optimal sequence of maintenance actions that minimize costs and economic risks over the structure life-cycle. By exploiting all model parameter distributions, the computed policy is not optimal only for specific parameters but accounts for all plausible values that the POMDP environment may assume. The resulting solution is thus robust to epistemic uncertainty over the model parameters. Only a few prior works have managed to combine Bayesian decision making and Dynamic Programming to obtain POMDP solutions that are robust to model uncertainty. In addition to the novel character of the formulation presented in this work, to the best of our knowledge this is also the first time that the two fields are involved with real-world application data.

This work opens up paths on both new applications and the development of methods for decision making under uncertainty. Possible extensions pertain to the hidden Markov model characteristics, used to infer the parameters of the POMDP environment, with time-dependent transition dynamics or hierarchical system dependencies comprising two possible further paths to explore. In the future, we further wish to investigate the use of Reinforcement Learning (RL) techniques for the development of solutions for maintenance planning that are robust to epistemic uncertainty, without any required prior knowledge of the problem. Several options can be explored along this path, such as the use of model-free \cite{zhu2017improving} or model-based \cite{arcieri2021model} algorithms, while multi-agent RL techniques can be merged with a hierarchical inferred model \cite{andriotis2019managing}. 

\section*{Acknowledgements}

The authors acknowledge the support of the Swiss Federal Railways (SBB) as part of the ETH Mobility Initiative project REASSESS.

\bibliography{bibliography}

\newpage

\appendix

\section{Computation of fractal values}
\label{app:fractal_values}

	\begin{algorithm}[H]
		\KwData{Longitudinal level band pass filtered to the range 1\,m to 70\,m}
	      \KwResult{Fractal values in short, mid and long wave range}	
		Definition of dividers $i$ $\epsilon$ $[5 \dots 580]$\;
		 \ForAll{dividers $i$}{
			Select $y$, 150\,m measurement window from longitudinal level signal\;
			Compute the polynomial length $L$ as:
		 $L(\lambda=\frac{150}{i})=\sum_{j=1}^{i}\sqrt{(x_j-x_{j-1})^{2}+(y(x_j)-y(x_{j-1}))^{2}}$
		 where $x$ and $y$ are the spacial window coordinates subdivided into $i$ segments\;
		Divide the Richardson plot into three sections (short, mid, long wave range) with delimiters: 
		\begin{itemize}
			\item{Delimiter section 1-2: $log(20’000\,mm/4)\simeq 3.7$}
			\item{Delimiter section 2-3: $log(3000\,mm/4)\simeq 2.9$}
		\end{itemize}	  
		\ForEach{section $s_i \in i={1,2,3}$}{
			Run a linear regression for $s_i$ on: $log(L(\lambda)) \ \forall \ log(\lambda) \in s_i$\;
			Return slope (corresponding to fractal value in wavelength bands $i$)\;}
		Repeat the fractal analysis taking a 150\,m signal window with a 1m shift\;
		}
		\caption{Computation of short, mid and long wave fractal values}
		\label{f:algoFraq}
	\end{algorithm}

\newpage

\section{Observation model parameters}
\label{app:inference_observation}

\begin{figure}[!htp]
    \centering
    
\begin{subfigure}[b]{1\textwidth}
\centering
\includegraphics[width=\linewidth]{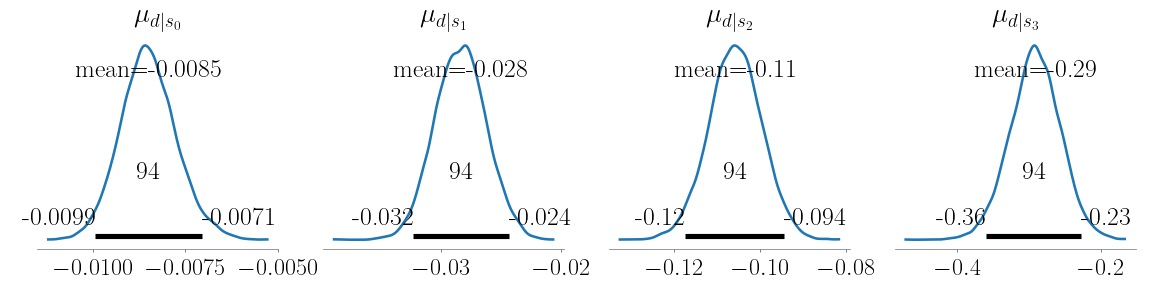}
\caption{Posterior distributions of state-dependent parameters $\mu_{d|s_t}$.}
\end{subfigure}
    
\begin{subfigure}[b]{1\textwidth}
\centering
\includegraphics[width=\linewidth]{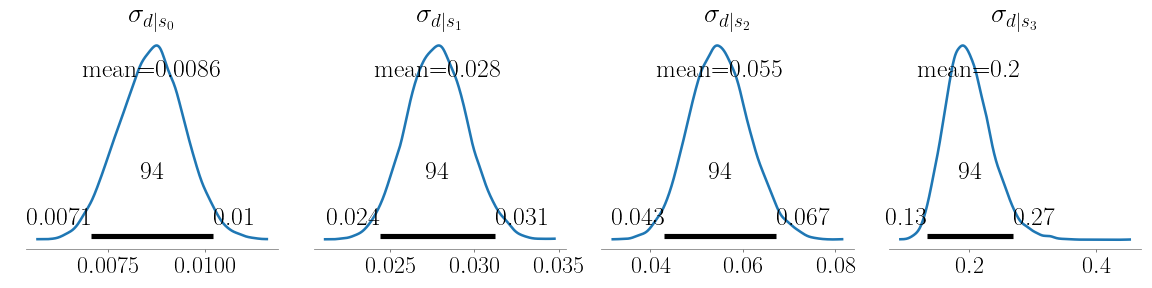}
\caption{Posterior distributions of state-dependent parameters $\sigma_{d|s_t}$.}
\end{subfigure}

\begin{subfigure}[b]{1\textwidth}
\centering
\includegraphics[width=\linewidth]{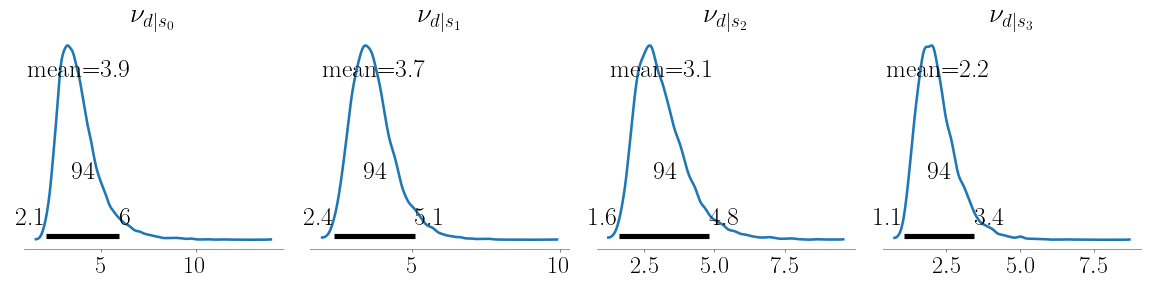}
\caption{Posterior distributions of state-dependent parameters $\nu_{d|s_t}$.}
\end{subfigure}

\caption{Posterior distributions of observation model parameters (deterioration process).}
        \label{fig:params_d}

\end{figure}

\begin{figure}[!htp]
    \centering

\begin{subfigure}[b]{1\textwidth}
\centering
\includegraphics[width=\linewidth]{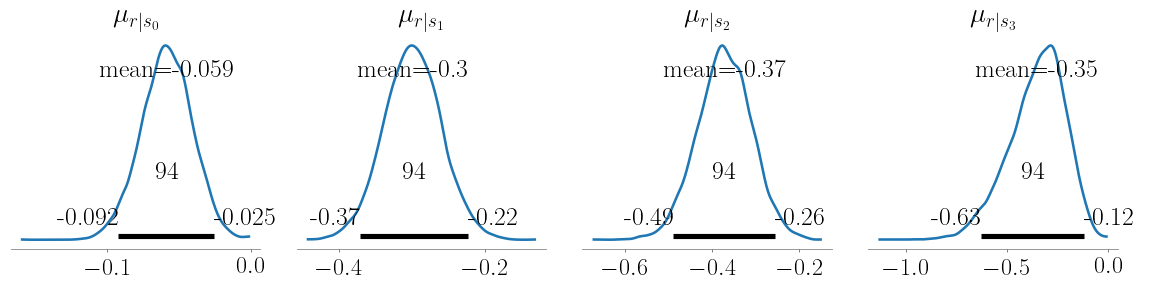}
\caption{Posterior distributions of state-dependent parameters $\mu_{r|s_t}$}
\end{subfigure}

\begin{subfigure}[b]{1\textwidth}
\centering
\includegraphics[width=\linewidth]{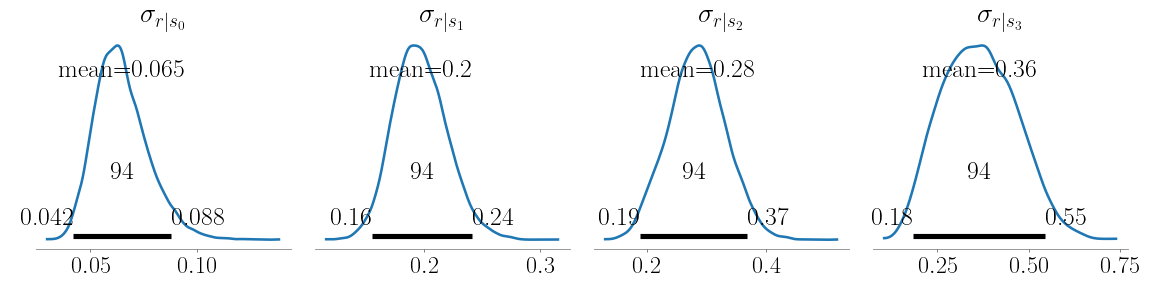}
\caption{Posterior distributions of state-dependent parameters $\sigma_{r|s_t}$.}
\end{subfigure}

\begin{subfigure}[b]{1\textwidth}
\centering
\includegraphics[width=\linewidth]{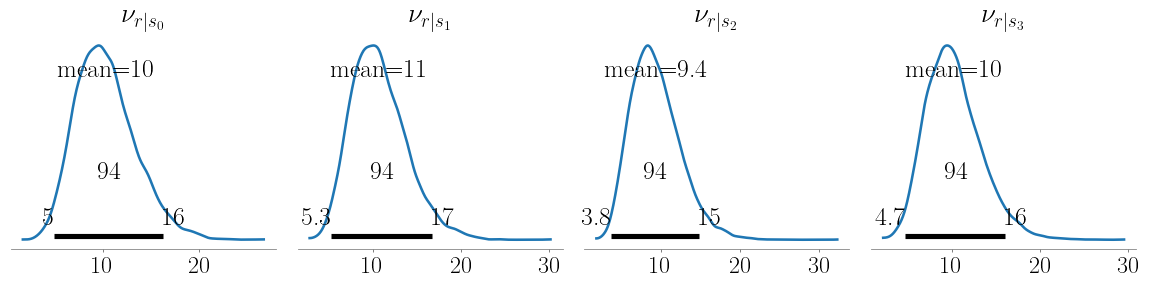}
\caption{Posterior distributions of state-dependent parameters $\nu_{r|s_t}$.}
\end{subfigure}

\begin{subfigure}[b]{1\textwidth}
\centering
\includegraphics[width=0.5\linewidth]{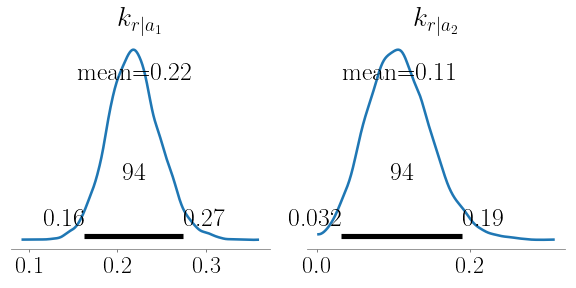}
\caption{Posterior distributions of the autoregressive parameters $k_{r|a_t}$ for $a_1$ (left) and $a_2$ (right).}
\end{subfigure}

\caption{Posterior distributions of observation model parameters (repair process).}
\label{fig:params_r}

\end{figure}

\begin{figure}[!htp]
    \centering

\begin{subfigure}[b]{1\textwidth}
\centering
\includegraphics[width=\linewidth]{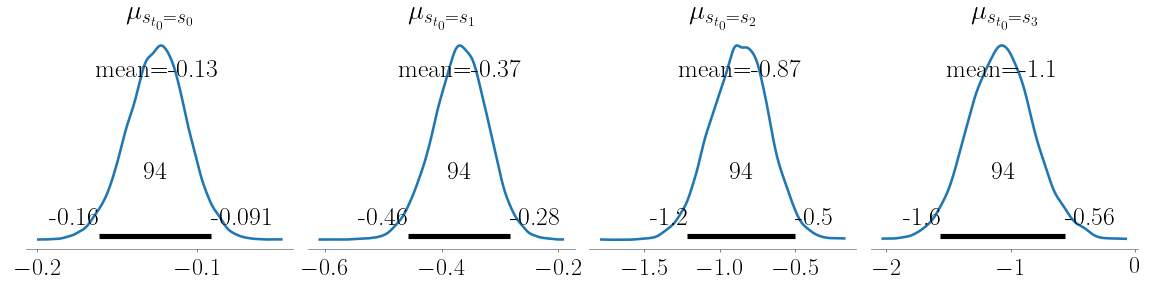}
\caption{Posterior distributions of parameters $\mu_{s_{t_0}}$.}
\end{subfigure}

\begin{subfigure}[b]{1\textwidth}
\centering
\includegraphics[width=\linewidth]{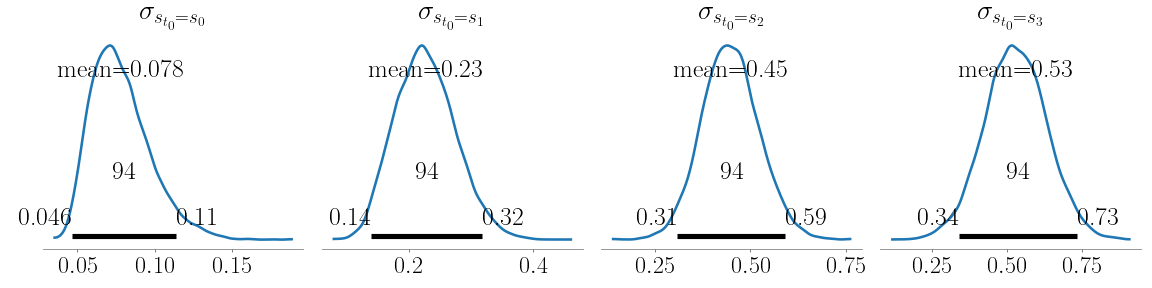}
\caption{Posterior distributions of parameters $\sigma_{s_{t_0}}$.}
\end{subfigure}

\begin{subfigure}[b]{1\textwidth}
\centering
\includegraphics[width=\linewidth]{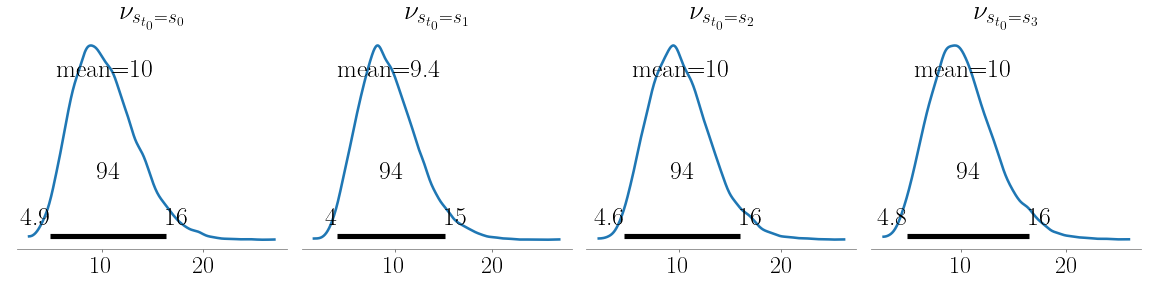}
\caption{Posterior distributions of parameters $\nu_{s_{t_0}}$.}
\end{subfigure}

\caption{Posterior distributions of observation model parameters (initial observation).}
\label{fig:params_i}
\end{figure}

\end{document}